\documentclass{article}

\usepackage{graphicx}
\usepackage{amssymb}
\usepackage{booktabs}
\usepackage{multirow}
\usepackage{tikz}
\usepackage{placeins}

\usepackage[utf8]{inputenc} 
\usepackage[T1]{fontenc}    
\usepackage{hyperref}       
\usepackage{url}            
\usepackage{booktabs}       
\usepackage{amsfonts}       
\usepackage{nicefrac}       
\usepackage{microtype}      
\usepackage{xcolor}         
\usepackage{float}
\usepackage[a4paper, margin=1in]{geometry}

\usepackage{graphicx}%
\usepackage{multirow}%
\usepackage{amsmath,amssymb,amsfonts}%
\usepackage{amsthm}%
\usepackage{mathrsfs}%
\usepackage[title]{appendix}%
\usepackage{xcolor}%
\usepackage{textcomp}%
\usepackage{manyfoot}%
\usepackage{booktabs}%
\usepackage{algorithm}%
\usepackage{algorithmicx}%
\usepackage{algpseudocode}%
\usepackage{listings}%

\usepackage{graphicx}
\usepackage{multirow}
\usepackage{hhline}
\usepackage{threeparttable}
\usepackage{adjustbox}

\usepackage{algpseudocode}
\definecolor{mynicegreen}{RGB}{0,200,0}
\usepackage{hhline}
\usepackage{adjustbox}
\usepackage{longtable}
\usepackage{tabularx}
\usepackage{csquotes}

\title{\textbf{Deep Transfer Learning Based Peer Review Aggregation and Meta-review Generation for Scientific Articles}}

\author{%
\normalsize
Md. Tarek Hasan$^{1}$, Mohammad Nazmush Shamael$^{1}$, H. M. Mutasim Billah$^{1}$, Arifa Akter$^{1}$\\
\normalsize
Md Al Emran Hossain$^{1}$, Sumayra Islam$^{1}$, Salekul Islam$^{2}$, Swakkhar Shatabda$^{3}$\\
\small
$^1$United International University \quad
$^2$North South University \quad
$^3$BRAC University \\
\small \texttt{tarek@cse.uiu.ac.bd}
}

\date{} 

\begin{document}

\maketitle

\begin{abstract}
\noindent
Peer review is the quality assessment of a manuscript by one or more peer experts. 
Papers are submitted by the authors to scientific venues, and these papers must be reviewed by peers or other authors. The meta-reviewers then gather the peer reviews, assess them, and create a meta-review and decision for each manuscript. As the number of papers submitted to these venues has grown in recent years, it becomes increasingly challenging for meta-reviewers to collect these peer evaluations on time while still maintaining the quality that is the primary goal of meta-review creation. In this paper, we address two peer review aggregation challenges a meta-reviewer faces: paper acceptance decision-making and meta-review generation. Firstly, we propose to automate the process of acceptance decision prediction by applying traditional machine learning algorithms. We use pre-trained word embedding techniques BERT to process the reviews written in natural language text. For the meta-review generation, we propose a transfer learning model based on the T5 model. Experimental results show that BERT is more effective than the other word embedding techniques, and the recommendation score is an important feature for the acceptance decision prediction. In addition, we figure out that fine-tuned T5 outperforms other inference models. Our proposed system takes peer reviews and other relevant features as input to produce a meta-review and make a judgment on whether or not the paper should be accepted. In addition, experimental results show that the acceptance decision prediction system of our task outperforms the existing models, and the meta-review generation task shows significantly improved scores compared to the existing models. For the statistical test, we utilize the Wilcoxon signed-rank test to assess whether there is a statistically significant improvement between paired observations.

\end{abstract}

\section{Introduction}
\label{sec1}
\begin{figure}[H]
    \centering
    \includegraphics[width = 0.75\textwidth]{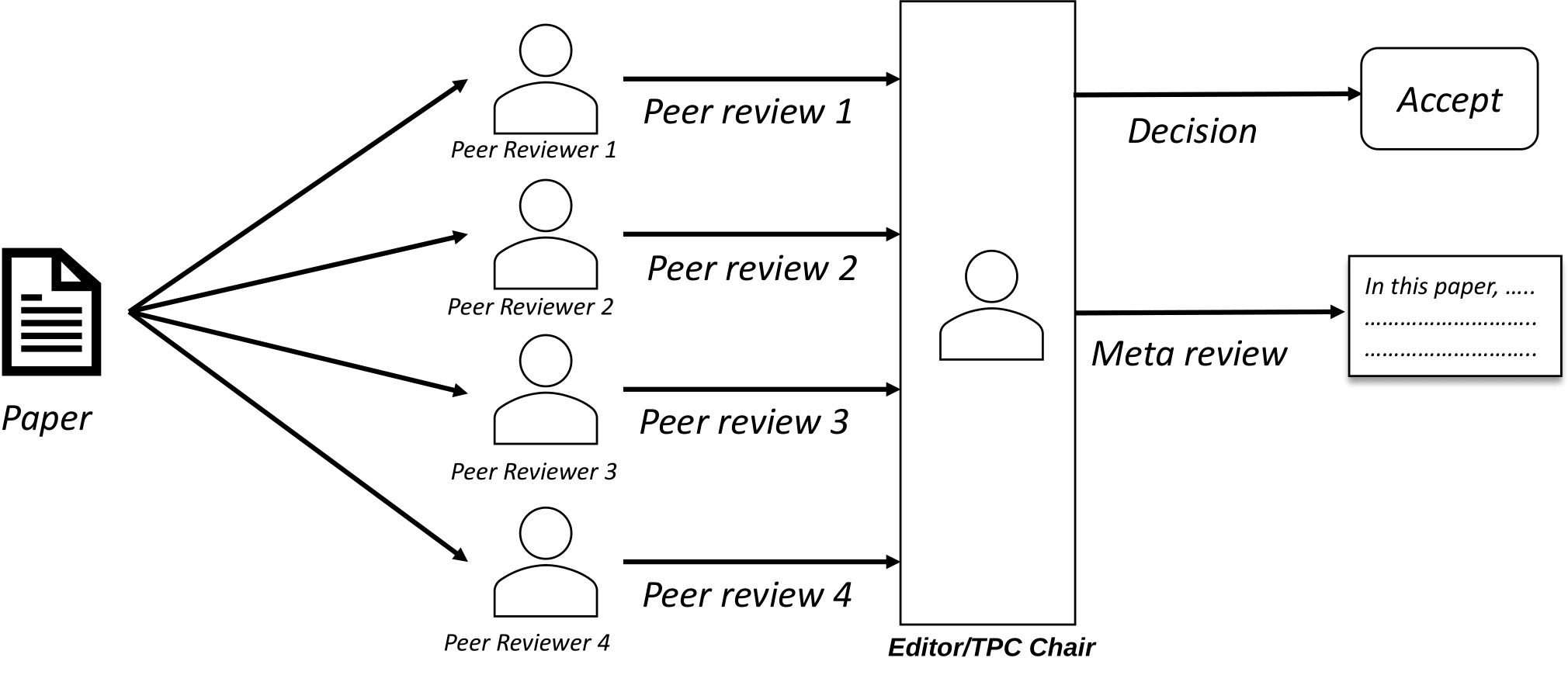}
    \caption{Process flow of meta-review generation}
    \label{fig:toy exmaple}
\end{figure}
Peer review is the process of evaluating a work by one or more persons who have similar skills as the creator of the work. For researchers, peer review is the most important foundation of trust. In academic or scientific conferences, papers are submitted by the authors and these papers need to be reviewed by peers or other authors as shown in Figure \ref{fig:toy exmaple}. After that, the meta-reviewers collect the peer reviews and go through the review. Finally, generate the meta-reviews along with the acceptance decision for each manuscript. As day by day, the number of papers submitted to these conferences is increasing rapidly, it is difficult for the Meta reviewers to aggregate these peer evaluations in a short time while maintaining the quality which is a vital role of meta-review generation. It is also difficult for them to decide whether the paper should be accepted or not. An automated tool might assist the meta-reviewer in completing this task in a very short time.\\
\noindent
Sometimes reviewers review context and their recommendations create some confusion for the meta-reviewer. For example, a reviewer's review context may be partially biased to the positive but his/her recommendation can be negative based on some points. If this similar case happened multiple times in a paper then the meta-reviewer might get confused. To avoid this confusion, our automated acceptance decision and meta-review generation will work very consistently and efficiently.\\
\noindent
Due to privacy issues with open peer reviews, the dataset was a huge challenge for our work. PeerRead \cite{kang2018dataset} is the widely used dataset for such types of work as acceptance decision prediction. However, the volume of data was insufficient to improve our work's accuracy. Additionally, there was no such dataset accessible for the generation of meta-reviews. We create datasets from OpenReview utilizing web scraping to generate meta-reviews. Additionally, we gather information from five international conferences. The next challenge was reformatting all the data in a set format after gathering the data from various sources.\\
\noindent
There are few works on acceptance decision prediction which are not foolproof. The previous works ignore necessary input features and use backdated overfitting prevention techniques. Some recent studies work on Transformer, a pre-trained language model that uses self-attention layers for learning. Some of the proposed text summarizers use BERT \cite{devlin2018bert}, which doesn't perform well in sequence-to-sequence tasks. UniLM \cite{dong2019unified} addresses this flaw of BERT \cite{devlin2018bert} by presenting a new training purpose for sequence-to-sequence tasks like text summarization. One of the most pertinent works \cite{pradhan2021deep} found on meta-review generation used UniLM \cite{dong2019unified} for sequence-to-sequence tasks but they have some false assumptions. Another issue of peer review-related tasks is the small size of publicly available data.\\
\noindent
We automated the process of creating meta-reviews and acceptance decision prediction in our project using machine learning. Our algorithm uses peer reviews and other features as input to produce a meta-review and make a judgment on whether the paper should be accepted or not. For acceptance decision prediction we utilize the paper title, abstract, peer reviews, recommendation, and confidence score as the feature. Meta-review is mainly an abstractive summary of all the peer reviews of that corresponding paper. Each sentence in the concatenated version of all peer reviews of a paper is classified as either positive, negative, or neutral based on polarity score, and then combined as pros, cons, and general information, creating a draft review that will pass to the model along with truncated meta-review for fine-tuning.\\
\noindent
In this paper, we gather datasets from different sources as there were relatively few data available for acceptance decision prediction and no datasets available for meta-review generation. Then we merge these datasets and create two different large datasets for two individual parts of our work. Then in the first part which is acceptance decision prediction, we use the most relevant features, paper title, abstract, peer review, recommendation score, and confidence score. We use traditional machine learning models for this part, and the experimental outcomes outperform all currently available models for the prediction of acceptance decisions. For the meta-review generation, we employ a transfer learning approach that is fine-tuned with the Text-To-Text Transfer Transformer (T5) \cite{raffel2019exploring} model. This T5 model is pre-trained on the C4 dataset and achieves a state-of-the-art result on many NLP benchmarks. Finally, for the evaluation of our work, we utilize accuracy and F1-score for acceptance decision prediction and ROUGE \cite{lin2004rouge} score for the meta-review generation and compare the result with the existing experimental results.

\section{Related Work}

Here, we have reported the existing works related to acceptance decision prediction and meta-review generation in separate subsections. In the following subsections, we have mentioned the used datasets, techniques, and their gaps for both tasks.

\subsection{Acceptance decision prediction}
Wang et al. \cite{wang2018sentiment} proposed a novel architecture to address the challenges with sentiment classification of peer reviews. The study analyzes the automation of sentiment classification of peer reviews which aims to classify the recommendation and acceptance sentiment of the review. The model classifies the sentences with positive and negative sentiments and uses that information to predict the acceptance decision. One shortcoming of the paper is that necessary features like paper title, recommendation score, and reviewers' confidence score are not considered for the model. The study uses Word2Vec which is no longer the state-of-the-art for sentiment analysis for Natural Language Understanding. Chakraborty et al. \cite{chakraborty2020aspect} used an active learning framework to build the training dataset for aspect prediction, which is used to capture the aspects and sentiments for the entire dataset. The study used aspect-based sentiment analysis on a dataset that doesn’t have all the aspects. So their accuracy didn't quite meet the desired standard. Ghosal et al. \cite{ghosal2019deepsentipeer} addressed the decision-making task of peer review. The proposed deep neural architecture used three information channels: the paper full-text, corresponding peer review texts, and sentiment. In the paper, the recommendation score is generated by using a regression model and decision based on the classification model. Kumar et al. \cite{kumar2022deepaspeer} devised a deep neural network architecture that aims to predict the acceptance and rejection decisions of the peer review process. They utilized three sources of information, peer review texts, associated aspects, and sentiment for predicting the acceptance decision. The experiments show the combination of aspect information (clarity, impact, originality, etc.) and sentiments (positive, negative) has a positive effect on the performance of the model.\\
\noindent
William et al. \cite{jen2018predicting} used the PeerRead \cite{kang2018dataset} dataset to produce a model to predict acceptance decisions. They worked on a Small-sized ICLR dataset that caused poor results with Neural Networks. Some features like ``abstract contain ML word”, ``title length”, ``average sentence length” etc. are used here which don't have much significance. The necessary features like recommendation score and confidence could be considered for better performance. For capturing information on abstract Word2vec is used but the current state-of-the-art is BERT. NLP techniques may be used to select features to describe the essence of paper content. Skorikov et al. \cite{skorikov2020machine} predicted the acceptance decision accurately for top-tier AI conferences like NIPS, ICLR, CONLL, ACL, and arXiv. An accuracy of 81\% is achieved using the Random Forest classifier. Bao et al. \cite{bao2021predicting} presented a framework to create decision sets for predicting paper acceptance, which can both accurately predict class labels and interpretably explain its decision boundaries. Wang et al. \cite{wang2021paper} predicted the acceptance rate of papers at the institution level. The paper formalizes the problem as a regression task that can rank the institution’s acceptance of articles on the objective conference. Wang et al. \cite{wang2020reviewrobot} presented the ReviewRobot for predicting review scores and generating detailed comments for every review category. They used unrelated features instead of significant features like recommendation scores and reviewer confidence scores. The one-hot-encoding technique is used for word embedding, but currently, state-of-the-art is BERT for Natural Language understanding. Bharti et al. \cite{bharti2023peerrec} propose a novel attention-based deep neural architecture that aims to predict the recommendation score and the acceptance decision. A sectional summary of the paper along with the associated review text and the reviewer's sentiment is used as input for the model. The study indicates that the inclusion of sentiment and attention has a positive effect on the final score of the model.
 
\subsection{Meta-review generation}
The purpose of Lou et al. \cite{louabstractive} is to construct a pipeline to cluster a vast volume of papers based on their scope and generate a brief human-readable summary of each cluster. In the article, the titles of all articles in one topic were concatenated to represent the corpus of the topic. They use the title, which is not the optimal representation of the content of scientific papers. On the other hand, after concatenating all the article titles, the highest token limit is up to 512 set by their used model. Esteva et al. \cite{esteva2020co} employs CAiRE, a multi-document summarization technique. CAiRE yields abstractive summaries by combining UniLM and BART, which is fine-tuned on a biomedical review dataset. They develop a summarizer that carries the retrieved documents and develops an abstractive summary until they reach an input length of 512 tokens. Kieuvongngam et al. \cite{kieuvongngam2020automatic} has worked on abstractive text summarization. They have used the COVID-19 Open Research Dataset. The keywords are taken from nouns and verbs only. But they should also add adjective parts to explore more accurately. Randomly dropping and adding words to the keyword sets can be used for more data augmentation. By doing this, more keyword summary pairs can be created. Liu et al. \cite{liu2018generative} proposed an adversarial process for abstractive text summarization. They first pre-train the generative model by generating summaries provided in the source text. Then pre-train the discriminator by giving positive samples from the human-generated summaries and negative samples from the pre-trained generator. Li et al. \cite{li2017deep} introduced a deep recurrent generative decoder to enhance the abstractive summarization performance. The model is a sequence-to-sequence oriented encoder-decoder framework provided with a latent system modeling element.\\
\noindent
Yuan et al. \cite{yuan2021can} design Natural Language Processing models to generate reviews for scientific papers that include more aspects of the paper than those created by humans. According to the assumption, their automatic peer review generation can be biased. They didn't provide any accept/reject decision or meta-review. Pradhan et al. \cite{pradhan2021deep} presented MetaGen They assumed that the reviewers update their reviews once the author addresses their comments. But this assumption is not always okay. Sometimes reviewers can also change their decision after seeing the new version of papers. The tokens are truncated and ignored in the training phase, where necessary tokens might have been ignored. Kang et al. \cite{kang2018a} introduced PeerRead \cite{kang2018dataset}, the first public peer-review dataset for research purposes. The paper presents prediction decisions and prophesies the numerical scores of review aspects. For high variance aspects like ‘originality’ and ‘impact’, simple models can outperform the mean baseline. Kumar et al. \cite{kumar2023towards} developed an approach to automatically generate decision-aware meta-reviews. The proposed model first predicts a recommendation score and an acceptance decision prediction. These scores are then used to generate the meta-reviews using a transformer-based seq2seq architecture. Experimental results show that including the sentiment of the review in the model yields better accuracy which in turn allows the model to generate higher-quality meta-reviews. Kumar et al. \cite{kumar2023deepmetagen} propose an unsupervised deep neural network approach for generating meta-reviews. They utilize an aspect-based sentiment analysis model to first deconstruct the review sentences based on their aspects and sentiments. Then a transformer model is used to reconstruct the reviews and filter opinions for summarization, resulting in concise meta-reviews that outperform competitive baselines.\\
\noindent
After analyzing the existing studies, we observe that very few papers are present on meta-review generation. Kousha et al. \cite{kousha2024artificial} reviews the existing literature related to partially or fully complete automation of publishing-related tasks. The paper discusses existing work on the automation of various aspects of the publishing cycle starting from journal suggestion systems for authors to post-publication peer review analysis, However, very little work is discussed regarding review decision prediction and automation of meta-review generation which falls in line with our analysis as well. All the current meta-review generators have some restrictions and there's room for improvement in terms of accuracy. On the other hand, the acceptance decision prediction also doesn't have any efficient solution. We are optimistic that this paper will make an advancement in predicting acceptance decisions along with qualitative meta-review generation. 

\begin{figure}[t]
    \makebox[\textwidth][c]{
    \includegraphics[width=1\textwidth]{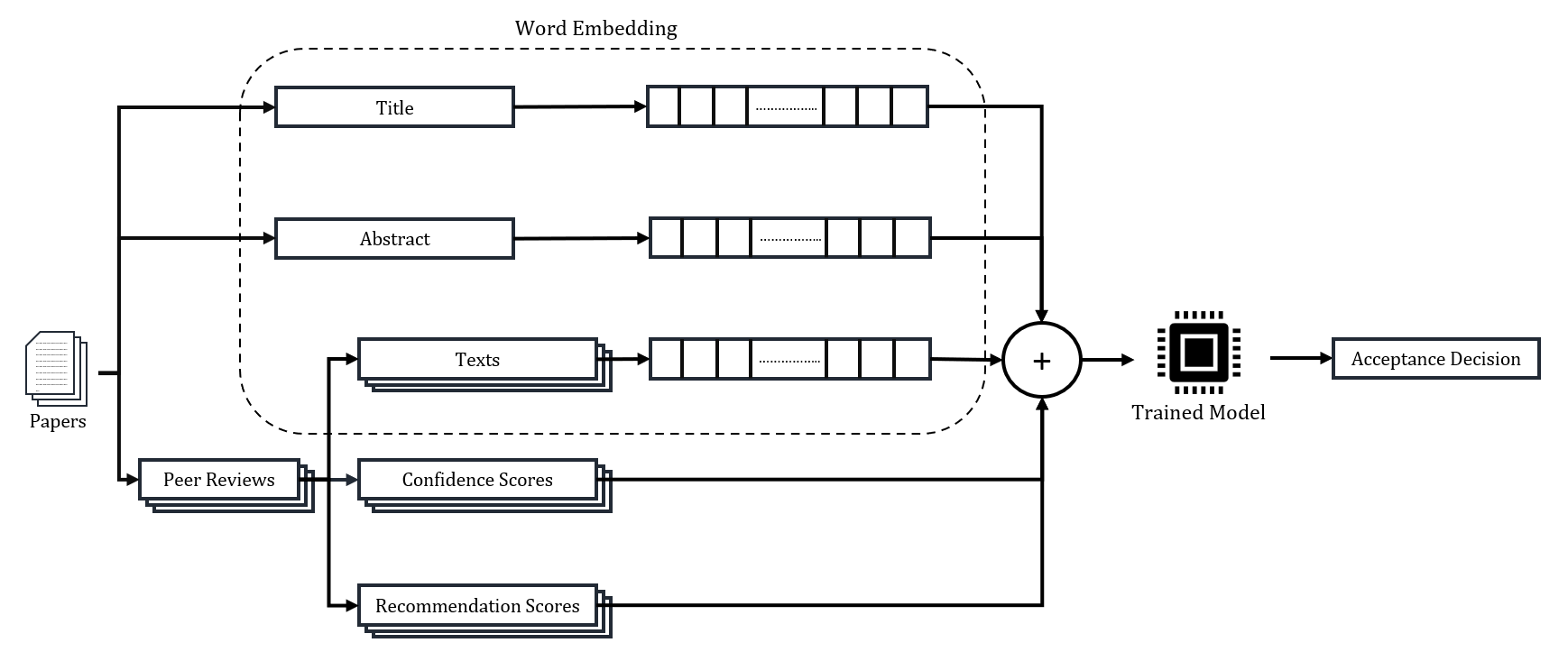}}
    \caption{Acceptance decision prediction process}
    \label{fig:AC decision prediction training}
\end{figure}

\section{Proposed method}
The process of meta-review generation is an essential part of any scientific or academic conference and journal. In this study, we propose a method that tries to simplify this process and make the job of the meta-reviewer easier. Shown in Figure \ref{fig:toy exmaple} is a user view of our work. Papers submitted by authors will be reviewed by peer reviewers. Based on their review our ML model will decide whether the paper should be accepted or not. It generates the meta-review as well. Due to the nature of the output provided by the ML model, the task can be broken down into two main components, acceptance decision prediction, and meta-review generation.

\begin{figure}
    \centering
    \includegraphics[width=1\textwidth]{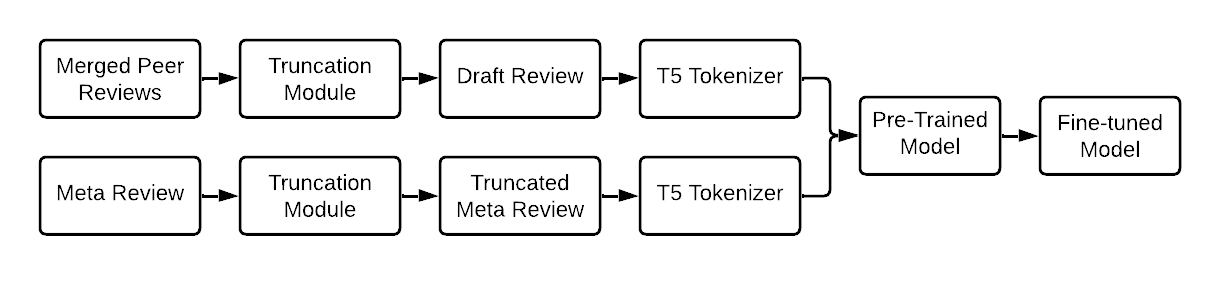}
    \caption{Process flow of T5 fine-tuning for meta-review generation}
    \label{fig:Process flow of t5 fine tuning}
\end{figure}

\begin{figure}
    \centering
    \includegraphics[width=0.80\textwidth]{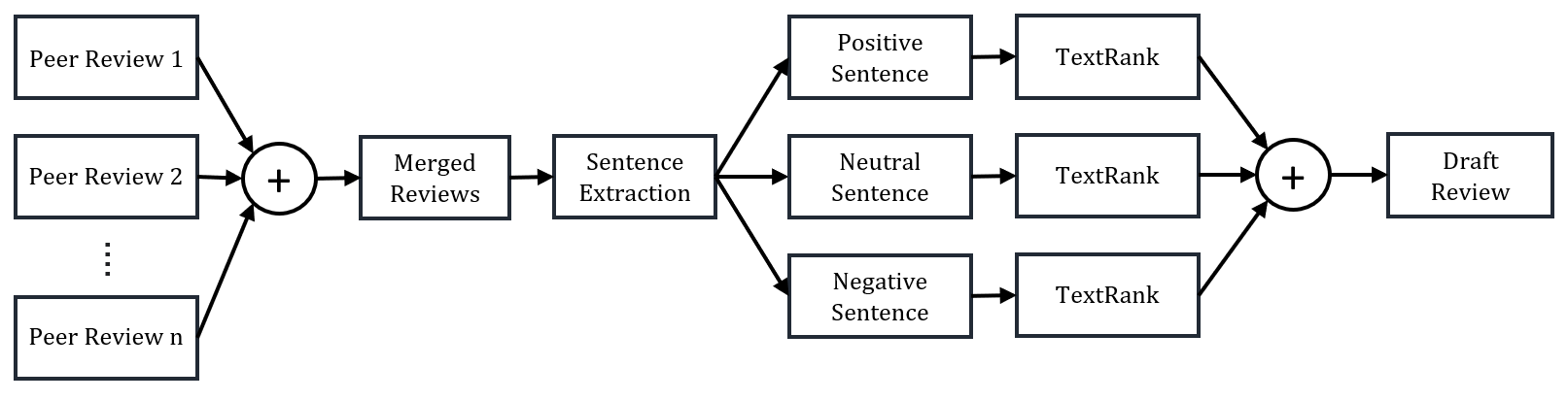}
    \caption{TextRank with Sentiment analysis}
    \label{fig:TextRank with Sentiment analysis}
\end{figure}

\subsection{Acceptance decision prediction} 
As we are working with Natural language, we need to first convert the natural language in a way to make it comprehensible to the computer. We used Bidirectional Encoder Representation from Transformers (BERT) \cite{devlin2018bert} as our preferred word embedding technique to do this task. BERT is based on transformer architecture and is very bi-directional. The model gathers information from both the left and right sides of a sentence’s context during the training phase. The secret to BERT’s success is its pre-trained step. Wikipedia (2500 million words) and Book Corpus (800 million words) are part of a vast corpus of unlabeled text that has been used to pre-train the BERT model. A model that has been trained with such a large number of texts has a far better knowledge of how a language functions than other word embedding techniques \cite{hasan2023review} like TF-IDF \cite{ramos2003using}, Word2Vec \cite{mikolov2013efficient} \cite{mikolov2013distributed}, GloVe \cite{pennington2014glove}, ELMo \cite{peters2018deep} and fastText \cite{joulin2016bag}.\\
Once the paper title, abstract, and peer reviews are converted using word embedding, the resulting vectors are concatenated with the recommendation score and confidence score to create the final feature set to train the model. For acceptance decision prediction traditional machine learning models like KNN \cite{fix1989discriminatory} \cite{altman1992introduction}, Decision Tree \cite{quinlan1987simplifying}, Random Forest \cite{breiman2001random}, Logistic Regression \cite{tranmer2008binary}, Naive Bayes \cite{huang2011naive}, and Support Vector Machine \cite{cortes1995support} algorithms are used. We have not used any deep learning techniques as we had favorable results with less complex traditional machine learning models with significantly less training time. The process of acceptance decision prediction is shown in Figure \ref{fig:AC decision prediction training}.\\

\renewcommand{\algorithmicrequire}{\textbf{Input:}}
\renewcommand{\algorithmicensure}{\textbf{Output:}}
\begin{algorithm}
\caption{Review Truncation using \textit{TextRank} and sentiment analysis}
\label{algo:draft_review}
\begin{algorithmic}[1]
\Require Review and word count
\Ensure Draft review
\Function{Summarize}{$text$, $word\_count$}
    \State $neutral \gets$ empty string
    \State $pros \gets$ empty string
    \State $cons \gets$ empty string
    \State $sentences \gets$ tokenize sentences($text$)
    \State $n \gets$ number of sentences
    \For{$i \gets 1$ to $n$}
        \State $sentence \gets$ sentences[i]
        \State $polarity\_score \gets$ sentiment\_analyzer($sentence$)
        \If{$polarity\_score = 0.0$}
            \State $neutral \gets neutral + sentence$
        \ElsIf{$polarity\_score > 0.0$}
            \State $pros \gets pros + sentence$
        \Else
            \State $cons \gets cons + sentence$
        \EndIf
    \EndFor
    \State $neutral \gets$ \textit{TextRank(neutral, word\_count = ceil(word\_count*0.2))}
    \State $pros \gets$ \textit{TextRank(pros, word\_count = ceil(word\_count*0.4))}
    \State $cons \gets$ \textit{TextRank(cons, word\_count = ceil(word\_count*0.4))}
    \State $draft\_review \gets neutral + pros + cons$
    \State \Return $draft\_review$
\EndFunction
\end{algorithmic}
\end{algorithm}

\subsection{Meta review generation} 
Text summarization can be of two types, extractive, and abstractive summarization. extractive summarization uses a subset of the sentences from the original text to create a summary whereas abstractive summarization uses contextual learning to recognize the text and creates its own words and phrases for creating a meaningful summary. For our work, we use a fine-tuned T5 model of Google for abstractive summarization of the peer reviews. Due to hardware limitations, we needed to limit review sizes by truncating tokens. We use two truncation techniques for truncating the tokens. TextRank \cite{mihalcea2004textrank} and TextRank with sentiment analysis. A simplified version of TextRank with sentiment analysis is shown in Figure \ref{fig:TextRank with Sentiment analysis}. For each sentence, we determine the sentiment of the sentence. The sentences with negative sentiment represent the cons and positive sentiment represents the pros of the paper. The sentences with neutral sentiments represent the general information of the paper. After the separation process of neutral, pros, and cons-related sentences, we use the TextRank algorithm for these three types of text. Finally, by concatenating the three parts, we get the draft review to send as the input of the model. After truncating the peer reviews and meta-reviews, we tokenize them using the T5 Tokenizer and use them to fine-tune the T5 model. Figure \ref{fig:Process flow of t5 fine tuning} shows a simplified version of the meta-review generation training process. The algorithmic details of review truncation using TextRank with sentiment analysis are shown in Algorithm \ref{algo:draft_review}.\\
\noindent
The algorithm \ref{algo:draft_review} takes the merged peer review and the number of words expecting the truncated peer review. The merged peer review means the concatenation of the peer reviews for each paper. Then, the algorithm extracts the sentences from the merged peer review. After that, for each sentence, it determines whether the sentence provides a positive sentiment or not based on the polarity score. If the polarity score is greater than zero, then the sentiment of the sentence is positive. In addition, the sentiment seems neutral if the polarity score is zero, and negative if the score is less than zero. The sentences with positive sentiments provide the pros of the paper, and the sentences with negative sentiments provide the cons of the paper. The algorithm concatenates the sentences with positive sentiment as pros, the sentences with negative sentiment as cons, and the rest of the sentences as neutral. Finally, to make the truncated review or draft review, the algorithms take 20\% of the total word count from neutral sentences, 40\% from pros, and 40\% from cons. Finally, by merging the truncated neutral, pros, and cons sentences, it generated the draft review.

\section{Experiment Analysis}
\subsection{Dataset Description}
Due to the work being separated into two different parts, two different datasets were used for the two experiments. With privacy concerns regarding public peer reviews, there are very few public peer review datasets available. For this reason, most datasets use publicly available open peer reviews available on websites like OpenReview\footnote{\href{https://openreview.net/}{https://openreview.net/}}. For the \textit{acceptance decision prediction}, we used one such dataset, the PeerRead \cite{kang2018dataset} dataset. For our experimental purposes, we used 19878 reviews from the PeerRead \cite{kang2018dataset} dataset. Among the 19878 reviews, there are 7270 reviews of ICLR 2017, 2854 reviews of ICLR 2018, 9440 reviews of ICLR 2019, 275 reviews of ACL 2017, and 39 reviews of CONLL 2017. 
For the \textit{\textbf{meta-review generation}}, a custom-tailored dataset was used for training and experimentation. we first gathered unstructured raw data from OpenReview where reviews and meta-reviews of major venues are openly available. Among the multiple venues it hosts, some of them contain a final meta-review provided by the program chair. We manually went through all the venues listed on their site and found seven venues containing meta-reviews. Among the seven venues, we found a total of 8955 papers of which 153 papers are from CORL 2021, 65 papers are from GI 2020, 491 papers are from ICLR 2017, 1059 papers are from ICLR 2018, 1579 papers are from ICLR 2019, 2594 papers are from ICLR 2020 and 3014 papers are from ICLR 2021 containing a meta-review provided by the program chair. The data was collected from the website using the OpenReview API which returns a JSON file containing various attributes of the review like title, abstract, keywords, code, comments, decision, meta-review, rating, confidence, direct replies, writer, etc. After gathering all the data and removing all features except peer reviews and meta-reviews and concatenating the peer reviews into one peer review, we combined the data to create the final dataset of 8955 meta-reviews and their corresponding peer reviews.
\subsection{Experimental Setup}
For the acceptance decision prediction, we use Google Colaboratory. Three types of runtime are provided by Google Colab. CPU, GPU, and TPU. We use GPU Tesla K80 with 12 GB of GDDR5 VRAM, an Intel Xeon Processor with two cores @ 2.20 GHz, and 13 GB RAM. For the meta-review generation, all the experiments are conducted on a 3.60GHz Intel(R) Core(TM) i7-7700 CPU, 32 GB RAM, and an NVIDIA TITAN XP with 12GB physical memory under Ubuntu 16.04.7 LTS. The reason behind using two different environments for the two different tasks is the time limit of Google Colaboratory as we need to execute the implementation for meta-review generation beyond the time limit of Google Colaboratory.
\subsection{Implementation Details}
\subsubsection{Acceptance Decision Prediction}
The features we get from the feature engineering phase are the paper title, paper abstract, peer reviews, recommendation score, and confidence score. Additionally, the class label is the acceptance decision. For the word embedding part of our work, we use BERT-base from BERTify\footnote{\href{https://github.com/khalidsaifullaah/BERTify}{BERTify}} and concatenated the last four layers of embedding which give the vector of size 764 * 4 = 3072 for the peer reviews of each paper. The same goes for each paper title and abstract of the paper. For the recommendation and confidence score, we just concatenate these with the vector of the paper title, abstract, and review. So, if we use the full feature set the size of the vector for each instance is 3072 + 3072 + 3072 + 8 + 8 = 9032 as we set the limit of the number of reviews for each paper as 8. For the experimental analysis, we used Word2Vec, and fastText as the word embedding techniques.\\
For implementing the word2Vec for sentences, we utilize Word2Vec from \textit{gensim} and the same vector size as BERT. For embedding the sentences in a text, we use the averaging technique after getting the word embedding for each of the words. Before passing the sentence in the embedding part, we preprocess the text i.e. Stemming, removing stopwords except for the word not as it is a very important word for the sentiment analysis task, and removing special characters from the text.\\
For implementing the fastText for sentences, we apply the same procedure as sentence embedding using Word2Vec. However, the only change is the vector size for each embedding because of the limitation of the resources. We use 2048 instead of 3072 here. As a result, the final vector size of each instance here is 2048 + 2048 + 2048 + 8 + 8 = 6160.\\
After getting the vectorized format for each instance, we utilize traditional machine learning algorithms from \textit{scikit-learn}. The algorithms we used are KNN (K = 5), Decision Tree, Random Forest, Logistic Regression, Naive Bayes, and SVM Linear. We use 80\% of the total data as training and 20\% for testing and spilt the dataset using \textit{train\_test\_split} from \textit{scikit-learn}.

\begin{table}[]
\centering
\caption{Instances of meta-reviews generated by T5, GPT2, Pegasus, BERT, BART.}

\begin{center}
\begin{scriptsize}
    \fcolorbox{black}{mynicegreen}{\rule{0pt}{2pt}\rule{2pt}{0pt}}  Positive sentiment 
\fcolorbox{black}{blue}{\rule{0pt}{2pt}\rule{2pt}{0pt}} Neutral sentiment
\fcolorbox{black}{red}{\rule{0pt}{2pt}\rule{2pt}{0pt}} Negative sentiment
\end{scriptsize}
\end{center}

\label{tab:examplesOfMetaReviewsAll}
\tiny
\begin{scriptsize}
\begin{tabular}{|p{2cm}|p{11cm}|}
\hline
\multicolumn{1}{|c|}{Actual Meta-Review}
&
{\color{blue}This paper studies the problem of Pareto fairness without having pre-defined protected groups.} {\color{mynicegreen} The reviewers agree that the problem studied here is interesting and relevant.} {\color{red} During the initial review period, reviewers identified a major correctness issue.} {\color{mynicegreen}The authors have then substantially changed the algorithm and experiments in the rebuttal period in order to address the issues.} {\color{blue} Now the convergence result in the paper follows more directly from the prior work of Chen et al.} {\color{red} Overall, the technical novelty of the paper appears to be limited.} {\color{blue} Even though the authors have also strengthened the related work discussion, they should also consider discussing the comparison between their work with that of Lahoti et al., as suggested by one of the reviewers.}

\\ \hline
\multicolumn{1}{|c|}{T5}
&
{\color{blue}Authors of a new algorithm for classifying at-risk groups need to elaborate. Authors use a multiplicative update on ``h" and project it back to the space of Pareto-efficient classifiers.} {\color{red} Authors say they are not sure if this trade-off is reasonable or whether it's significant.} {\color{red} If it's true, it's not a good idea to apply the same algorithm to a classifier, says Adam McCarthy.}
\\ \hline
\multicolumn{1}{|c|}{GPT2}
&
{\color{red} It is unclear whether this bound is reasonable or whether the loss in efficiency is significant.} {\color{blue}They provide a one-line proof stating ``Immediate application of Theorem 11 in Chen et al. Was the idea to perform a multiplicative update on ``h" and project it back to the space of Pareto-efficient classifiers? Instead, why is the MW update applied again to the alphas?} {\color{blue} At the very least, I think authors need to elaborate on how exactly they use the algorithm from Chen et al. Why are subgroups based on outcomes relevant? Why would it make sense in any application to consider based on outcomes? In comparison to Duchi et.al 2020, and Hashimoto et a. They aim to provide an algorithm that outputs a classifier that minimizes the risk of any group (of sufficient size) while at the same time being Pareto efficient. This section also gives us an idea about the trade-off between efficiency and fairness. The authors show that the solution to this problem degenerates into a trivial classifier when the size of the smallest group is below a certain threshold.} {\color{mynicegreen} I like the problem formulation and find the problem setup to be practically relevant (given that there's an increasing emphasis in the fairness community on addressing scenarios where the protected group information is noisy or unknown).} {\color{red} However, I'm unable to verify the correctness of the algorithm and guarantee- The authors say they adapt Algorithm 3 from Chen et al.}
\\ \hline
\multicolumn{1}{|c|}{Pegasus}
&
{\color{blue} There has been a lot of discussion about how to improve the accuracy of a classifier when it is trying to trade off the accuracy of an off group against the accuracy of an off group.}
\\ \hline
\multicolumn{1}{|c|}{BART}
&
 {\color{blue} The authors aim to provide an algorithm that minimizes the risk of any group (of sufficient size) while at the same time being Pareto efficient. They do this by first showing that this problem is equivalent to the case when there are only two groups and then using no-regret dynamics.} {\color{mynicegreen} The improvement over the current state of the art in this problem (DRO) is very marginal and in some cases (the Compas Dataset), DRO does slightly better than BPF.}
\\ \hline
\multicolumn{1}{|c|}{BERT}
&
{\color{red} It is unclear whether this bound is reasonable or whether the loss in efficiency is significant.} {\color{blue} Was the idea to perform a multiplicative update on "h" and project it back to the space of Pareto-efficient classifiers? Instead, why is the MW update applied again to the alphas?} {\color{red} At the very least, I think authors need to elaborate on how exactly they use the algorithm from Chen et al.} {\color{blue} The current discussion is light on the details. To obtain a minimax classifier, the authors deploy a multiplicative weight update-based method, which is somewhat standard.} {\color{mynicegreen} Pros: BPF addresses fairness beyond demographics, that is, it does not rely on predefined notions of at-risk groups, neither at train nor at test time.} {\color{blue} Why are subgroups based on outcomes relevant? Why would it make sense in any application to consider based on outcomes?} {\color{red} Weaknesses:* The improvement over the current state of the art in this problem (DRO) is very marginal and in some cases (the Compas Dataset), DRO does slightly better than BPF.} {\color{blue} It would help to discuss it for the readers who have not read Martinez et al (2020) After the Rebuttal I thank the authors for their clarifications and efforts to improve their work.} {\color{mynicegreen} POST-REVISION: Thanks for revising your algorithm and clarifying its theoretical properties.} {\color{mynicegreen} Experimental results show that the proposed framework improves worst-case risk in multiple standard datasets, while simultaneously providing better levels of service for the remaining population, in comparison to competing methods.}
\\ \hline
\end{tabular}
\end{scriptsize}

\end{table}

\subsubsection{Meta-review Generation}
The feature we get from the feature engineering phase is peer reviews and the output for this part is meta-reviews. We use PyTorch to fine-tune the T5 base of Google. Because of the limitation of computational resources, we need to truncate the peer reviews as well as the meta-reviews. As \textit{T5TokenizerFast} just takes the first \textit{n} tokens and deducts the rest of the tokens where \textit{n} is a user-defined argument, there is a high chance of missing some important information. That is why we use the \textit{TextRank} algorithm to extract the most informative sentences from the text. In addition, for the experimental analysis, we use the sentiment analysis technique along with \textit{TextRank} algorithm for the peer reviews as shown in Algorithm \ref{algo:draft_review} for this purpose. We utilize \textit{SentimentIntensityAnalyzer} from \textit{nltk} for the sentiment analysis, and the \textit{TextRank} \cite{mihalcea2004textrank} algorithms we used the \textit{summarizer} from \textit{gensim}. We set the token size for peer review to 300, and for meta-review 150. Here, we use 80\% of the total dataset for the training purpose and the rest of the data for the test. We use 20\% of the training data for validation purposes. We utilize Adam as an optimizer with batch size 8 and a learning rate of 0.001. To store the best model, we use a callback that monitors the validation loss. Moreover, we train the whole T5 base for 5 epochs.\\
For the experimental analysis, we used inference model BERT \cite{devlin2018bert}, GPT2-large \cite{radford2019language}, BART-large \cite{lewis2019bart}, Pegasus \cite{zhang2020pegasus} fine-tuned of xsum, T5-base \cite{raffel2019exploring}. We used BERT and GPT2-large from \textit{summarizer} and the rest of the inference model from Hugging Face\footnote{\href{https://huggingface.co/}{Hugging Face}}. As the inference models also have limitations \cite{louabstractive}, we set the limit of token size for the peer reviews to 512, and the generated meta-review to 250. An example of the generated meta-review of different inference models can be seen in Table\ref{tab:examplesOfMetaReviewsAll}. The generated summary of the T5 fine-tuned model is shown in Table \ref{tab:examplesOfMetaReviewsT} and \ref{tab:examplesOfMetaReviewsTSA}.

\begin{table}[]
\centering
\caption{Instances of T5 fine-tuned model generated meta-reviews (Truncation Technique during fine-tuning: TextRank)}

\begin{center}
\begin{scriptsize}
    \fcolorbox{black}{mynicegreen}{\rule{0pt}{2pt}\rule{2pt}{0pt}}  Positive sentiment 
\fcolorbox{black}{blue}{\rule{0pt}{2pt}\rule{2pt}{0pt}} Neutral sentiment
\fcolorbox{black}{red}{\rule{0pt}{2pt}\rule{2pt}{0pt}} Negative sentiment
\end{scriptsize}
\end{center}

\label{tab:examplesOfMetaReviewsT}
\begin{scriptsize}

\begin{tabular}{|p{7cm}|p{7cm}|}
\hline
\multicolumn{1}{|c|}{Actual meta-review} &
  \multicolumn{1}{c|}{T5 fine-tuned model generated meta-review} \\ \hline
{\color{blue}This paper studies the problem of Pareto fairness without having pre-defined protected groups.} {\color{mynicegreen}The reviewers agree that the problem studied here is interesting and relevant.} {\color{red} During the initial review period, reviewers identified a major correctness issue.} {\color{mynicegreen} The authors have then substantially changed the algorithm and experiments in the rebuttal period to address the issues.} {\color{mynicegreen} Now the convergence result in the paper follows more directly from the prior work of Chen et al.} {\color{red} Overall, the technical novelty of the paper appears to be limited.} {\color{blue} Even though the authors have also strengthened the related work discussion, they should also consider discussing the comparison between their work with that of the authors, as suggested by one of the reviewers.} &
 {\color{blue} This paper proposes a method for learning a Pareto-efficient classifier that minimizes the risk of any group (of sufficient size) while at the same time being Pareto-efficient.} {\color{mynicegreen} The reviewers agree that the problem is interesting, and the results are promising.} {\color{red} However, they also raised concerns about the experimental setup and the empirical evaluation.} {\color{red} In particular, it seems that the proposed algorithm does not outperform the state-of-the-art in terms of efficiency and performance.} \\ \hline
{\color{blue} The paper proposes a hierarchical diversity-promoting regularizer for neural networks.} {\color{blue} Experiments are shown with this regularizer applied to the last fully connected layer of the network, in addition to L2 and energy regularizers on other layers.} {\color{red} Reviewers found the paper well-motivated but had concerns about the writing/readability of the paper and that it provides only marginal improvements over existing simple regularizers such as L2.} {\color{blue} I would encourage the authors to look for scenarios where the proposed regularizer can show clear improvements and resubmit to a future venue.} &
 {\color{blue} This paper proposes hierarchical hyperspherical regularization of neural networks.} {\color{mynicegreen} The reviewers found that the paper is well-written and easy to follow.}  {\color{red} However, there were concerns about the novelty of the approach, as well as the lack of comparison with related work (Liu et al. 2018) and the experimental evaluation.} {\color{red} In particular, it was not clear why the proposed approach would be better than other regularizers such as Liu et al. (2018).} \\ \hline
{\color{blue} The paper considers the question of identifying whether a model is bad from an OOD perspective or certifying that it is good.} {\color{red} The reviews agree that there are interesting ideas in the paper, however, a lack of sufficient experiments and presentation issues were pointed out which make the paper not ready for acceptance at this stage.} &
{\color{blue}  This paper proposes a new metric to measure the variance in the model when the data from each environment is unweighted.} {\color{red} The reviewers agree that the proposed metric is interesting and useful, but there are concerns about the novelty of the method as well as the experimental results.} {\color{red} In particular, the authors claim that they do not provide any empirical evidence that the proposed metric can be generalized to OOD samples.} {\color{red} However, the reviewers also point out that the experiments don't seem convincing enough to support the claims.} \\ \hline
\end{tabular}
\end{scriptsize}

\end{table}

\begin{table}[]
\centering
\caption{Instances of T5 fine-tuned model generated meta-reviews (Truncation Technique during fine-tuning: TextRank with Sentiment Analysis)}

\begin{center}
\begin{scriptsize}
    {\fcolorbox{mynicegreen}{mynicegreen}{\rule{0pt}{2pt}\rule{2pt}{0pt}}  Positive sentiment}
\fcolorbox{blue}{blue}{\rule{0pt}{2pt}\rule{2pt}{0pt}} Neutral sentiment
\fcolorbox{red}{red}{\rule{0pt}{2pt}\rule{2pt}{0pt}} Negative sentiment
\end{scriptsize}
\end{center}

\label{tab:examplesOfMetaReviewsTSA}
\begin{scriptsize}

\begin{tabular}{|p{7cm}|p{7cm}|}
\hline
\multicolumn{1}{|c|}{Actual meta-review} &
  \multicolumn{1}{c|}{T5 fine-tuned model generated meta-review} \\ \hline
{\color{blue}This paper studies the problem of Pareto fairness without having pre-defined protected groups.} {\color{mynicegreen}The reviewers agree that the problem studied here is interesting and relevant.} {\color{red} During the initial review period, reviewers identified a major correctness issue.} {\color{mynicegreen} The authors have then substantially changed the algorithm and experiments in the rebuttal period in order to address the issues.} {\color{mynicegreen} Now the convergence result in the paper follows more directly from the prior work of Chen et al.} {\color{red} Overall, the technical novelty of the paper appears to be limited.} {\color{blue} Even though the authors have also strengthened the related work discussion, they should also consider discussing the comparison between their work with that of the authors, as suggested by one of the reviewers.} &
  {\color{blue} This paper proposes a new optimization algorithm for Pareto-efficient classifiers.} {\color{mynicegreen}The reviewers agree that the idea is interesting, and the results are promising.} {\color{red} However, there are some concerns about the significance of the proposed approach, as well as the clarity of the presentation.} \\ \hline
{\color{blue} The paper proposes a hierarchical diversity-promoting regularizer for neural networks.} {\color{blue} Experiments are shown with this regularizer applied to the last fully connected layer of the network, in addition to L2 and energy regularizers on other layers.} {\color{red} Reviewers found the paper well-motivated but had concerns about the writing/readability of the paper and that it provides only marginal improvements over existing simple regularizers such as L2.} {\color{blue} I would encourage the authors to look for scenarios where the proposed regularizer can show clear improvements and resubmit to a future venue.} &
 {\color{blue} This paper proposes a hierarchical learning approach for the problem of diversity in neural networks.} {\color{mynicegreen} The reviewers agree that the idea is interesting, and the results are promising.} {\color{red} However, there are concerns about the significance of the proposed approach, as well as the quality of the empirical evaluation.} \\ \hline
{\color{mynicegreen} The paper considers the question of identifying whether a model is bad from an OOD perspective or certifying that it is good.} {\color{red}The reviews agree that there are interesting ideas in the paper, however, a lack of sufficient experiments and presentation issues were pointed out which make the paper not ready for acceptance at this stage.} &
 {\color{blue} This paper proposes a method for training OOD models to detect whether an OOD model fails.} {\color{mynicegreen} The reviewers agree that the paper is well-written and easy to follow.} {\color{red}However, there are some concerns about the clarity of the presentation and the experimental results.} {\color{red}In particular, it is not clear how the proposed approach differs from existing approaches.} \\ \hline
\end{tabular}
\end{scriptsize}

\end{table}

\subsection{Evaluation Metrics}
For the testing of the acceptance decision prediction model, we utilize \textit{accuracy\_score}, and \textit{f1\_score} from \textit{scikit-learn}. For evaluating the meta-review generation, we use Recall-Oriented Understudy for Gisting Evaluation (ROUGE) score \cite{lin2004rouge} from PyPI \footnote{\href{https://pypi.org/project/rouge-score}{ROUGE score}}, and we report the f-measure for ROUGE-1, ROUGE-2, ROUGE-L, and ROUGE-Lsum as the mean of all meta-reviews. We use stemming while evaluating the generated meta-reviews.\\

ROUGE is a metric to automatically measures the degree of similarity by comparing machine-generated summaries with human-created (reference) summaries. The primary concept is to allocate a single numerical score to a summary that tells how good it is compared to one or more reference summaries. 

\paragraph{\textbf{ROUGE-N:}} ROUGE-N calculates the number of matching n-grams between the model-generated summary and the reference. In equation \ref{eqn:rouge_n}, the ROUGE-N formula \cite{tay2019red} is shown. Here gram-n is the
selection of n-gram and S is the human-generated reference summary:
\begin{equation}
\label{eqn:rouge_n}
R O U G E-N=\frac{\Sigma g r a m_{n \in S} \mathrm{Count}_{m a t c h}\left(\mathrm{gram}_n\right)}{\Sigma g r a m_{n \in S} \mathrm{Count}\left(\mathrm{gram}_n\right)}
\end{equation}
\noindent 
ROUGE-1  compares the uni-grams between the model-generated summary and the reference summary. The ROUGE-1 metric has separate precision and recall.
ROUGE-1 precision is calculated as the ratio of the number of uni-grams in the model-generated summary that also appears in the reference summary, over the number of uni-grams in the model generated.
ROUGE-1 recall is calculated as the ratio of the number of uni-grams in reference summary that also appears in machine-generated summary, over the number of uni-grams in reference.
\\
ROUGE-2 uses bi-grams instead of uni-grams. ROUGE-2 recall indicates the ratio of the number of matches in machine-generated and the reference summary. On the other hand, the ROUGE-2 precision is the ratio of the number of bi-grams matches and the number of bi-grams in the machine-generated summary. 
\paragraph{\textbf{ROUGE-L:}} 
Instead of comparing n-grams, ROUGE-L \cite{lin2004automatic} \cite{sanchez2020experimental} is a ratio
between the length of the machine-generated summaries’ longest common sub-sequence (LCS) and the length of the reference. The idea is that the longer the LCS of two summaries is, the more similar the two summaries are. One of the benefits is that LCS doesn't need consecutive positions within the reference sequences. It calculates sentence-level word ordering as n-grams.

\paragraph{\textbf{ROUGE-Lsum:}} 
Summary-level LCS \cite{lin2004rouge} computes the Union and compares the reference summary sentence, and every model-generated summary sentence.

\begin{table}[]
\centering
\caption{Summary of ablation study for acceptance decision prediction}
\label{tab:subset_of_features}
\begin{scriptsize}
\begin{tabular}{|c|c|c|c|c|c|}
\hline
\multirow{2}{*}{Experiment} & \multicolumn{5}{c|}{Features} \\ \hhline{|~|-----|} 
 & Paper Title & Abstract & Peer Review & Recommendation Score & Confidence Score \\ \hline
1 & $\sqrt{}$ & $\sqrt{}$ & $\sqrt{}$ & $\sqrt{}$ & $\sqrt{}$ \\ \hline
2 & $\sqrt{}$ & $\sqrt{}$ & $\sqrt{}$ & $\sqrt{}$ & \\ \hline
3 & $\sqrt{}$ & $\sqrt{}$ & $\sqrt{}$ & & \\ \hline
4 & & & $\sqrt{}$ & $\sqrt{}$ & $\sqrt{}$ \\ \hline
5 & & & $\sqrt{}$ & $\sqrt{}$ & \\ \hline
6 & & & $\sqrt{}$ & & \\ \hline
7 & $\sqrt{}$ & & $\sqrt{}$ & $\sqrt{}$ & $\sqrt{}$ \\ \hline
\end{tabular}
\end{scriptsize}
\end{table}

\begin{table}[]
\centering
\caption{Results of Ablation Study 1 for acceptance decision prediction using PeerRead \cite{kang2018dataset} only}
\label{tab:experiment 1}
\scriptsize
\begin{tabular}{|c|c|*{6}{c|}}
\hline

\multicolumn{1}{|l|}{Exp.} & \multicolumn{7}{c|}{Experimental   Results} \\ \hline
\multirow{9}{*}{1} & \multirow{3}{*}{Algorithm} & \multicolumn{6}{c|}{Word Embedding Technique} \\  \hhline{~~------} 
 &  & \multicolumn{2}{c|}{BERT} & \multicolumn{2}{c|}{Word2Vec} & \multicolumn{2}{c|}{fastText} \\ \hhline{~~------} 
 &  & \textbf{ACC (\%)} & \textbf{F1 (\%)} & \textbf{ACC (\%)} & \textbf{F1 (\%)} & \textbf{ACC (\%)} & \textbf{F1 (\%)} \\ \hhline{~-------} 
 & KNN (K = 5) & 68.07 & 38.38 & 81.47 & 71.66 & 71.15 & 48.92 \\ \hhline{~-------} 
 & Decision Tree & \textbf{83.39} & \textbf{74.39} & 80.59 & 71.17 & 78.67 & 67.72 \\ \hhline{~-------} 
 & Random Forest & 68.88 & 42.58 & 67.13 & 24.8 & 69.06 & 27.76 \\ \hhline{~-------} 
 & Logistic Regression & 80.94 & 70.93 & 83.04 & 73.71 & 80.24 & 69.38 \\ \hhline{~-------} 
 & Naive Bayes & 52.62 & 44.12 & 49.13 & 44.78 & 47.55 & 44.24 \\ \hhline{~-------} 
 & SVM Linear & 81.82 & 71.89 & 81.64 & 71.54 & 82.17 & 71.98 \\ \hline

\multirow{9}{*}{2} & \multirow{3}{*}{Algorithm} & \multicolumn{6}{c|}{Word Embedding Technique} \\ \hhline{~~------} 
 &  & \multicolumn{2}{c|}{BERT} & \multicolumn{2}{c|}{Word2Vec} & \multicolumn{2}{c|}{fastText} \\ \hhline{~~------} 
 &  & \textbf{ACC (\%)} & \textbf{F1 (\%)} & \textbf{ACC (\%)} & \textbf{F1 (\%)} & \textbf{ACC (\%)} & \textbf{F1 (\%)} \\ \hhline{~-------} 
 & KNN (K = 5) & 63.46 & 33.65 & 70.1 & 52.89 & 68.88 & 47.65 \\ \hhline{~-------} 
 & Decision Tree & 67.66 & 52.69 & 63.99 & 47.72 & 62.41 & 46.38 \\ \hhline{~-------} 
 & Random Forest & 65.73 & 40.24 & 67.13 & 27.69 & 68.01 & 24.07 \\ \hhline{~-------} 
 & Logistic Regression & 74.3 & 53.33 & 74.65 & 54.26 & 72.55 & 50.47 \\ \hhline{~-------} 
 & Naive Bayes & 52.27 & 43.71 & 48.95 & 44.7 & 47.72 & 44.53 \\ \hhline{~-------} 
 & SVM Linear & 75.17 & 52.03 & \textbf{75.87} & 54.3 & \textbf{75.87} & 54.61 \\ \hline

\multirow{9}{*}{3} & \multirow{3}{*}{Algorithm} & \multicolumn{6}{c|}{Word Embedding Technique} \\ \hhline{~~------} 
 &  & \multicolumn{2}{c|}{BERT} & \multicolumn{2}{c|}{Word2Vec} & \multicolumn{2}{c|}{fastText} \\ \hhline{~~------} 
 &  & \textbf{ACC (\%)} & \textbf{F1 (\%)} & \textbf{ACC (\%)} & \textbf{F1 (\%)} & \textbf{ACC (\%)} & \textbf{F1 (\%)} \\ \hhline{~-------} 
 & KNN (K = 5) & 58.92 & 31.88 & 58.04 & 30.64 & 59.27 & 34.37 \\ \hhline{~-------} 
 & Decision Tree & 59.44 & 26.11 & 58.22 & 39.49 & 58.39 & 41.67 \\ \hhline{~-------} 
 & Random Forest & 59.44 & 24.68 & 62.94 & 13.11 & 66.26 & 21.22 \\ \hhline{~-------} 
 & Logistic Regression & 65.73 & 11.71 & 67.83 & 30.3 & \textbf{68.18} & 35.92 \\ \hhline{~-------} 
 & Naive Bayes & 52.27 & 43.71 & 48.95 & \textbf{44.69} & 47.03 & 43.78 \\ \hhline{~-------} 
 & SVM Linear & 65.21 & 3.86 & 65.38 & 0 & 66.43 & 10.28 \\ \hline

\multirow{9}{*}{4} & \multirow{3}{*}{Algorithm} & \multicolumn{6}{c|}{Word Embedding Technique} \\ \hhline{~~------} 
 &  & \multicolumn{2}{c|}{BERT} & \multicolumn{2}{c|}{Word2Vec} & \multicolumn{2}{c|}{fastText} \\ \hhline{~~------} 
 &  & \textbf{ACC (\%)} & \textbf{F1 (\%)} & \textbf{ACC (\%)} & \textbf{F1 (\%)} & \textbf{ACC (\%)} & \textbf{F1 (\%)} \\ \hhline{~-------} 
 & KNN (K = 5) & 84.79 & \textbf{77.37} & 82.87 & 73.08 & 74.48 & 55.76 \\ \hhline{~-------} 
 & Decision Tree & 83.92 & 75.79 & 81.29 & 72.91 & 79.37 & 69.59 \\ \hhline{~-------} 
 & Random Forest & \textbf{84.97} & 77.04 & 72.2 & 41.33 & 73.95 & 46.59 \\ \hhline{~-------} 
 & Logistic Regression & 82.34 & 72.33 & 82.52 & 72.83 & 83.74 & 75.33 \\ \hhline{~-------} 
 & Naive Bayes & 61.01 & 24.92 & 45.8 & 45.61 & 42.48 & 46.33 \\ \hhline{~-------} 
 & SVM Linear & 82.52 & 72.83 & 82.69 & 73.32 & 82.34 & 72.33 \\ \hline

\multirow{9}{*}{5} & \multirow{3}{*}{Algorithm} & \multicolumn{6}{c|}{Word Embedding Technique} \\ \hhline{~~------} 
 &  & \multicolumn{2}{c|}{BERT} & \multicolumn{2}{c|}{Word2Vec} & \multicolumn{2}{c|}{fastText} \\ \hhline{~~------} 
 &  & \textbf{ACC (\%)} & \textbf{F1 (\%)} & \textbf{ACC (\%)} & \textbf{F1 (\%)} & \textbf{ACC (\%)} & \textbf{F1 (\%)} \\ \hhline{~-------} 
 & KNN (K = 5) & 70.8 & \textbf{55.46} & 70.8 & 53.22 & 67.66 & 45.1 \\ \hhline{~-------} 
 & Decision Tree & 72.73 & 51.85 & 65.91 & 49.61 & 66.78 & 51.78 \\ \hhline{~-------} 
 & Random Forest & 74.13 & 54.6 & 66.26 & 28.25 & 66.61 & 30.54 \\ \hhline{~-------} 
 & Logistic Regression & 75.17 & 52.67 & 74.48 & 52.9 & 72.73 & 51.55 \\ \hhline{~-------} 
 & Naive Bayes & 60.67 & 23.73 & 45.63 & 45.53 & 41.96 & 45.75 \\ \hhline{~-------} 
 & SVM Linear & 74.65 & 47.65 & 76.05 & 54.79 & \textbf{76.22} & 55.26 \\ \hline

\multirow{9}{*}{6} & \multirow{3}{*}{Algorithm} & \multicolumn{6}{c|}{Word Embedding Technique} \\ \hhline{~~------} 
 &  & \multicolumn{2}{c|}{BERT} & \multicolumn{2}{c|}{Word2Vec} & \multicolumn{2}{c|}{fastText} \\ \hhline{~~------} 
 &  & \textbf{ACC (\%)} & \textbf{F1 (\%)} & \textbf{ACC (\%)} & \textbf{F1 (\%)} & \textbf{ACC (\%)} & \textbf{F1 (\%)} \\ \hhline{~-------} 
 & KNN (K = 5) & 64.86 & 4.74 & 56.99 & 28.48 & 60.49 & 35.43 \\ \hhline{~-------} 
 & Decision Tree & 65.21 & 3.86 & 59.62 & 44.07 & 56.99 & 39.11 \\ \hhline{~-------} 
 & Random Forest & 65.21 & 3.86 & 65.56 & 25.66 & 62.41 & 21.82 \\ \hhline{~-------} 
 & Logistic Regression & 65.21 & 3.86 & 67.48 & 29.007 & \textbf{67.66} & 34.16 \\ \hhline{~-------} 
 & Naive Bayes & 60.49 & 23.13 & 45.63 & 45.53 & 41.96 & \textbf{45.93} \\ \hhline{~-------} 
 & SVM Linear & 65.21 & 3.86 & 65.38 & 0 & 65.38 & 0 \\ \hline

\multirow{9}{*}{7} & \multirow{3}{*}{Algorithm} & \multicolumn{6}{c|}{Word Embedding Technique} \\ \hhline{~~------} 
 & & \multicolumn{2}{c|}{BERT} & \multicolumn{2}{c|}{Word2Vec} & \multicolumn{2}{c|}{fastText} \\ \hhline{~~------} 
 & & \textbf{ACC (\%)} & \textbf{F1 (\%)} & \textbf{ACC (\%)} & \textbf{F1 (\%)} & \textbf{ACC (\%)} & \textbf{F1 (\%)} \\ \hhline{~-------} 
 & KNN (K = 5) & 77.97 & 63.79 & 82.87 & 73.22 & 74.13 & 54.6 \\ \hhline{~-------} 
 & Decision Tree & \textbf{87.62} & \textbf{76.96} & 83.74 & 75.59 & 80.77 & 71.05 \\ \hhline{~-------} 
 & Random Forest & 78.32 & 64.97 & 68.88 & 33.58 & 70.45 & 37.17 \\ \hhline{~-------} 
 & Logistic Regression & 81.82 & 71.43 & 82.17 & 72.28 & 82.87 & 73.66 \\ \hhline{~-------} 
 & Naive Bayes & 53.32 & 43.55 & 47.03 & 45.8 & 42.13 & 42.63 \\ \hhline{~-------} 
 & SVM Linear & 81.99 & 71.63 & 83.04 & 73.85 & 81.99 & 71.78 \\ \hline
\end{tabular}
\end{table}

\subsection{Results and Discussion}
\subsubsection{Acceptance Decision Prediction}
To predict the acceptance decision for the manuscript, we perform the ablation study. We conduct seven experimental analysis using different subsets of features from the feature set. The summary of the ablation study is shown in Table \ref{tab:subset_of_features}. We add and remove features from the feature set for the experiments to figure out the features that create more values for the model for predicting acceptance decision prediction. In the first experiment of the ablation study, we keep all the features for the experiment. In the next experiment of the ablation study, we remove the confidence score to check whether the confidence score creates any biases or not for predicting the acceptance of a manuscript. In the third experiment, we remove the recommendation score and the confidence score to determine the importance of the recommendation score and confidence score for this problem. In the next experiments, we keep and remove the paper title and abstract to figure out whether these features play any vital role in predicting acceptance decisions.\\
\noindent 
\textit{Classifier Selection}\\
We use traditional machine learning algorithms KNN \cite{fix1989discriminatory} \cite{altman1992introduction}, Decision Tree \cite{quinlan1987simplifying}, Random Forest \cite{breiman2001random}, Logistic Regression \cite{tranmer2008binary}, Naive Bayes \cite{huang2011naive} and Support Vector Machine \cite{cortes1995support} for the experimental analysis. To find the best-fit machine learning algorithm for the acceptance decision prediction, we conduct the ablation study on the PeerRead \cite{kang2018dataset} dataset as it is a widely used dataset for the acceptance decision prediction. We name this ablation study 1 where we perform seven experimental analysis. In each experiment, we use the subset of features to figure out the importance of each feature. 80\% of the total dataset is used for the training purpose and the rest of the dataset is used for testing. The experimental results in Table \ref{tab:experiment 1} show that we get the best results in terms of accuracy in experiment 4 where we do not use the paper title and abstract as the feature. This represents that the paper title and abstract are not that important for the acceptance decision prediction. For each experiment, the best results are marked in bold in Table \ref{tab:experiment 1}.\\
\noindent
We get the maximum accuracy of 84.94\% and a competitive F1 Score of 77.04\% using the Random Forest algorithm, as shown in experiment 4 in Table \ref{tab:experiment 1}. In addition, we use BERT as the word embedding technique here. In experiments 1 and 7, the best-performing algorithm is the Decision Tree. In experiments 2 and 5, the best-performing algorithm is the SVM Linear, and in experiments 3 and 6, it is Logistic Regression. The performance of Logistic Regression over all the experiments is consistent as we can see in the table that the second-best result for experiments 1, 2, 5, and 7 have been obtained by using Logistic Regression as the classifier. In addition, another important observation from the experimental results is the performance of Random Forest is not consistent over the experiments rather the performance is very poor in other experiments. We choose three consistent performing models Decision Tree, Logistic Regression, and SVM Linear to compare the performance.\\ 
\begin{figure}
    \centering
    \includegraphics[width=0.90\textwidth]{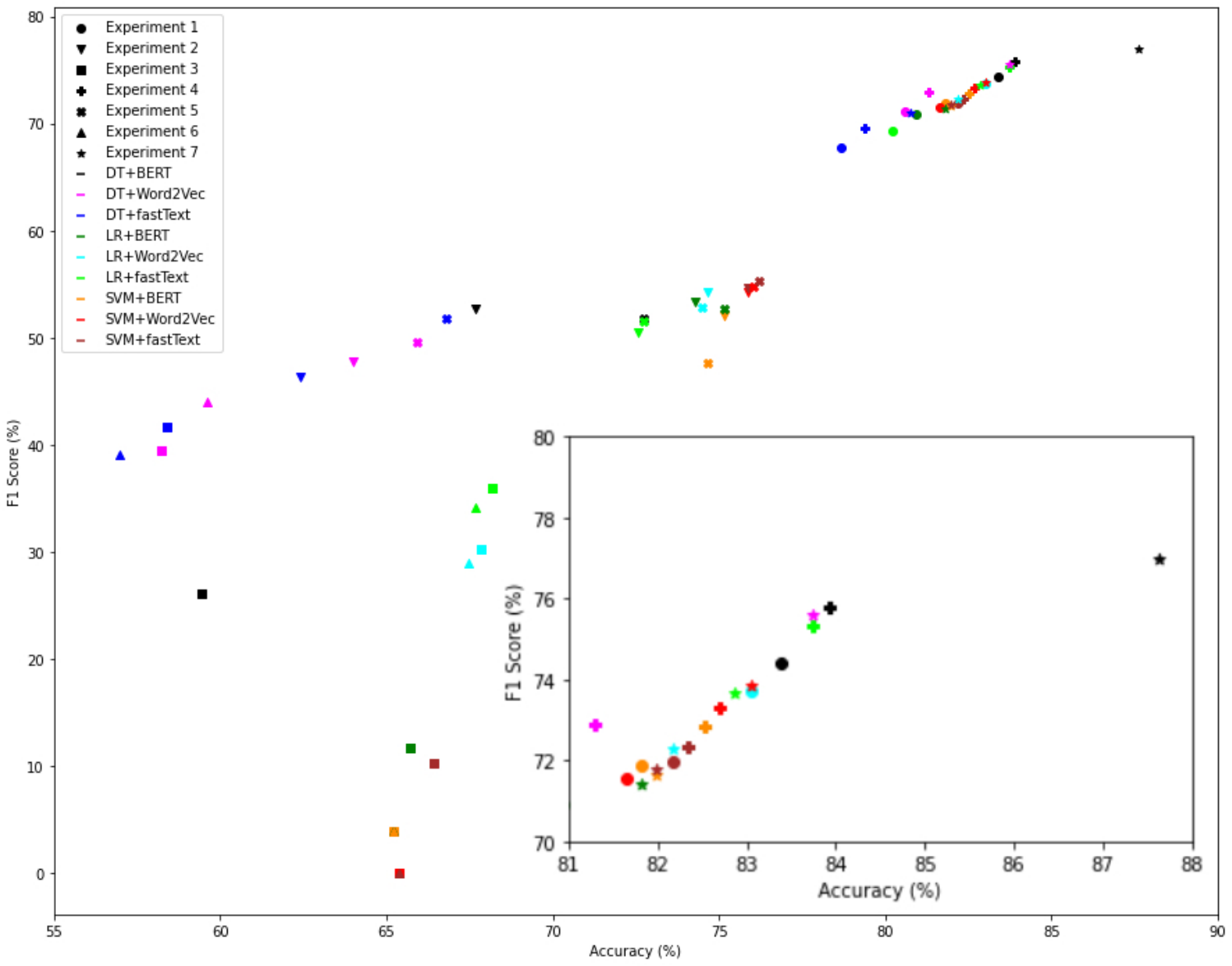}
    \caption{Performance Comparison of different ML algorithms on PeerRead \cite{kang2018dataset} Dataset}
    \label{fig:Experiment1}
\end{figure}
\noindent
As we can see in the figure \ref{fig:Experiment1}, the Decision Tree is the best-performing algorithm among the other consistent algorithms on the PeerRead \cite{kang2018dataset} dataset. In addition to that, we can say that BERT is the best word embedding technique among BERT, Word2Vec, and fastText. We get the best results in terms of accuracy and F1 Score using Decision Tree as the classifier and BERT as the word embedding technique in Experiment 7. The best accuracy and the F1 score we get are in order 87.62\% and 76.96\%. In experiment 7, we used Paper Title, Peer Review, Recommendation Score, and Confidence Score as the features. So, we can conclude by saying that the abstract of the paper is not that important to decide whether the paper should be accepted or not for the machine learning-based classifier. From the experimental analysis, we can state that the Decision Tree as the classifier and BERT as the word embedding technique are producing consistent results. For the statistical proof of this statement, we conduct the Wilcoxon signed-rank test \cite{woolson2007wilcoxon} based on accuracy score and F1-Score and figure out that, the performance of this combination is significant compared to all other combinations. While performing all the statistical tests in this work, the value of alpha is set to 0.05.\\
\noindent 
\textit{Experiments with other datasets}\\
We conduct the experiments on two other datasets. So, we use PeerRead \cite{kang2018dataset} for training and \textit{Conference [1, 5]} for testing as the second ablation study. If we carefully notice the results of ablation study 2 in Table \ref{tab:experiment 2}, the selected classifier Decision Tree and the word embedding technique BERT perform poorly on experiment 7. We observe from the experimental results when we use confidence as the feature in experiments 1, 4, and 7, the performance of the selected combination falls drastically. This represents that the confidence score is creating noise as the feature and the reviewers of the \textit{Conference [1,5]} were not sincere while giving the confidence scores. But in the third ablation study, we figure out that when we train the model using a merged dataset of PeerRead \cite{kang2018dataset} and \textit{Conference [1,5]}, then the model performs much better than the ablation study 2 as per the result analysis of Table \ref{tab:experiment 3}. In addition to that, we get the best results in terms of accuracy and F1 Score of the selected combination (Decision Tree and BERT) in order 86.92\% and 80.58\% in ablation study 3 whereas in ablation study 2 the accuracy and F1 Scores are in order 45.4\% and 41.6\%. From the experimental analysis, we can come to a decision that, we need to follow proper rules and regulations while giving the review, recommendation score, and confidence score, so that we do not need to train the model using datasets of different conferences to predict the acceptance decision accurately. We can create a benchmark dataset and by training the model using that dataset, we will be able to predict the acceptance decision for the manuscripts of all the conferences which will save valuable time.

\subsubsection{Meta-review Generation}

For the meta-review generation, we use various inference models and the T5 transformer to generate the meta-review. These inference models summarize the input text while capturing the important aspects of the input text. Due to computational limitations, it is necessary to shorten both the peer and meta-reviews. For the inference models, the token size of peer reviews is set to 512 and meta-reviews to 250. However, the T5 transformer model demands more computational resources and thus, the token size for peer reviews is reduced to 300 and for meta-reviews to 150. Our work is divided into two ablations studies based on the truncation technique applied in the time of fine-tuning T5. The peer reviews are truncated in two ways, one of the truncation techniques is the standard TextRank \cite{mihalcea2004textrank} algorithm and another one is the TextRank with sentiment analysis.\\
\begin{table}
\centering
\caption{Comparison between TextRank and TextRank with sentiment analysis}

\begin{center}
\begin{scriptsize}

\fcolorbox{mynicegreen}{mynicegreen}{\rule{0pt}{2pt}\rule{2pt}{0pt}} Positive sentiment
\fcolorbox{blue}{blue}{\rule{0pt}{2pt}\rule{2pt}{0pt}} Neutral sentiment
\fcolorbox{red}{red}{\rule{0pt}{2pt}\rule{2pt}{0pt}} Negative sentiment
\end{scriptsize}
\end{center}

\label{tab:TextRankExample}
\begin{scriptsize}
\begin{tabular}{|p{3cm}|p{11cm}|}
\hline
     Merged Review
     &  
     {\color{blue} This paper presents an annotation tool to assist in the labeling of language or vision datasets as used for machine learning. The claimed contribution of this tool are modularity and, in a way, interoperability.} {\color{blue} While I can see the merits of a unified tool, I am not sure what is proposed in this paper qualifies as a novel contribution to the field of HCI.} {\color{red}The tool may indeed be useful, but the only claim to this comes from its use for some of the authors' own projects -- while this is presented instead as a universal tool addressing the issue of different research labs using different tools.} {\color{blue}Given this claim, one would expect some external validation, grounded in HCI methodology. Secondly, I am not sure why there is a problem in the first place. Surely over the past many decades, if indeed the lack of a standardized tool is such a dire situation, there would be attempts at creating such a solution.} {\color{red} The motivation for this being a problem is not argued convincingly enough in the paper.} {\color{blue} One possible approach to this would be again grounded in HCI methodology, such as conducting interviews with current users of annotation tools. Finally, in my almost 30 years of experience, I have used various tools for these purposes, and none of these was proprietary to the labs where I conducted research. Some annotation tools were provided open source or free from various universities or institutes -- this suggests that there aren't any insurmountable barriers to sharing (and potentially, standardizing) such tools.} {\color{red} Given this, it is not clear to me whether the proposed interface solves a real problem.} {\color{mynicegreen} This doesn't mean that the proposed tool is not good, but it means that its value proposition may not be the one claimed by the authors in the abstract.} {\color{blue} One minor point that I would encourage the authors to consider in a subsequent revision of the paper: the ``integration of language and vision in machine learning applications" is not necessarily new, and more importantly, it is the other way around, especially for language (it is recently that ML has started being the de facto approach to language processing).} {\color{mynicegreen} Overall, in my view, this paper may be of interest to the community in terms of becoming aware of a potentially useful tool. }{\color{red}However, there is not enough in terms of research contribution for a full-length paper (a demo or poster may be more suitable).} {\color{mynicegreen}I would like to thank the authors for submitting this work to GI2020. The paper describes a comprehensive data annotation tool that has components and features for language, vision, and relations labeling that are customizable and switchable to suit various labeling needs.} \\\hline
     TextRank
     & 
     {\color{blue} This paper presents an annotation tool to assist in the labeling of language or vision datasets as used for machine learning.} {\color{blue} While I can see the merits of a unified tool, I am not sure what is proposed in this paper qualifies as a novel contribution to the field of HCI.} {\color{red} The tool may indeed be useful, but the only claim to this comes from its use for some of the authors' projects -- while this is presented instead as a universal tool addressing the issue of different research labs using different tools.} {\color{blue} Given this claim, one would expect some external validation, grounded in HCI methodology.} {\color{red} The motivation for this being a problem is not argued convincingly enough in the paper.} {\color{blue} One possible approach to this would be again grounded in HCI methodology, such as conducting interviews with current users of annotation tools.} {\color{red} Given this, it is not clear to me whether the proposed interface solves a real problem.} {\color{red} However, there is not enough in terms of research contribution for a full-length paper (a demo or poster may be more suitable).}\\\hline
     TextRank with Sentiment Analysis
     & 
    {\color{blue} The claimed contribution of this tool is modularity and, in a way, interoperability. Given this claim, one would expect some external validation, grounded in HCI methodology. One possible approach to this would be again grounded in HCI methodology, such as conducting interviews with current users of annotation tools.  Finally, in my almost 30 years of experience, I have used various tools for these purposes, and none of these was proprietary to the labs where I conducted research. This doesn't mean that the proposed tool is not good, but it means that its value proposition may not be the one claimed by the authors in the abstract.} {\color{mynicegreen} The paper describes a comprehensive data annotation tool that has components and features for language, vision, and relations labeling that are customizable and switchable to suit various labeling needs. This paper presents an annotation tool to assist in the labeling of language or vision datasets as used for machine learning.} {\color{red} The motivation for this being a problem is not argued convincingly enough in the paper. Given this, it is not clear to me whether the proposed interface solves a real problem. The tool may indeed be useful, but the only claim to this comes from its use for some of the authors' projects -- while this is presented instead as a universal tool addressing the issue of different research labs using different tools. However, there is not enough in terms of research contribution for a full-length paper (a demo or poster may be more suitable).}\\\hline
\end{tabular}
\end{scriptsize}
\end{table}
\noindent
The TextRank algorithm is an algorithm that can be used to find the most relevant sentences in a text. A graph is constructed with sentences as vertex and their similarity as their edges. Based on this graph, the sentences are sorted using a ranking algorithm. This way we can take the most important sentences based on the rank of the sentence and create a summary of the text. TextRank with sentiment analysis takes this a step further and first separates the sentences into three groups based on whether or not the sentence represents a positive point, a negative point, or a neutral point. Then TextRank is applied to these groups and the most relevant sentences of each group are concatenated to create the summary of the text. In this phase, we take 20\% of neutral sentences from all the neural sentences, 40\% positive points and the rest are negative points. We decided to take 20\%-40\%-40\% of neutral, positive, and negative points to make a balanced review so that we can get an unbiased review to make a proper meta-review.  As an illustration of these truncation techniques, let us consider the examples given in Table \ref{tab:TextRankExample}.\\
\noindent
As we can see in Table \ref{tab:TextRankExample}, the TextRank and TextRank with sentiment analysis outputs provide different perspectives on the peer review of the paper. The TextRank algorithm focuses on the major points mentioned by the reviewers, and it carries the same sentiment throughout the whole review. On the contrary, the TextRank with Sentiment analysis focuses on the negative aspects along with the positive sides mentioned by the reviewers. In addition, it also shows some neutral points about the paper. So, we can say that TextRank algorithms are biased rather than TextRank with sentiment analysis compared to the output of each algorithm for a given peer review. Moreover, due to the concatenation of the different aspects of the review for the summary when using TextRank with sentiment analysis, the resulting summary comes out neutral. Because of this behavior, TextRank with sentiment analysis cannot properly summarize the text as overly positive or overly negative. As a result, TextRank proves to be more effective for truncation compared to TextRank with sentiment analysis.\\
\begin{table}[]
\centering
\caption{Summary of ablation study for meta-review generation}
\label{tab:ablation2}
\begin{tabular}{|c|cc|}
\hline
  & \multicolumn{2}{c|}{Truncation}                                          \\ \hhline{|~|--|}
\multirow{-2}{*}{Experiment} & \multicolumn{1}{l|}{Generated}                & \multicolumn{1}{l|}{Reference} \\ \hline
1 & \multicolumn{1}{c|}{}                         &                          \\ \hline
2 & \multicolumn{1}{c|}{}                         & {$\sqrt{}$} \\ \hline
3 & \multicolumn{1}{c|}{{$\sqrt{}$}} &                          \\ \hline
4                            & \multicolumn{1}{c|}{{$\sqrt{}$}} & {$\sqrt{}$}       \\ \hline
\end{tabular}
\end{table}
\noindent
We need to truncate the peer review as well as the meta-review during the phase of fine-tuning because of the limitation of the resources. We conduct two ablation studies during the training phase based on the truncation technique being used. For ablation study 1, we truncate the merged peer reviews with TextRank, and for ablation study 2, we truncate the merged peer review with TextRank with sentiment analysis. As for the Meta-Review, in both ablation studies, we use TextRank for truncation.\\
We use the truncated peer review as input for the inference models. Examples of the generated meta-review of different inference models and the fine-tuned T5 model can be seen in Table \ref{tab:examplesOfMetaReviewsAll}. In general, GPT2 and BERT produced larger summaries than the summaries of BART and T5. On the contrary, Pegasus made the meta-review too concise and the review lost a lot of its meaning. As demonstrated by Table \ref{tab:examplesOfMetaReviewsT} and Table  \ref{tab:examplesOfMetaReviewsTSA}, the fine-tuned T5 generated the most realistic meta-reviews when compared to the inference models.  Furthermore, the structure of the meta-reviews generated by the T5 model was found to be similar to that of actual meta-reviews, whereas the other inference models sometimes produced meta-reviews with inconsistent structures.\\
As we use TextRank to truncate the meta-review and two different truncation techniques to truncate the peer reviews during the training phase, to evaluate the performance of the model we use the four methods illustrated in Table \ref{tab:ablation2}. 
\begin{displayquote}
\textbf{Experiment-1:} Neither the actual nor generated meta-reviews will be truncated.

\textbf{Experiment-2:} The generated meta-reviews will remain unchanged, but the actual meta-reviews will be truncated.

\textbf{Experiment-3:} The actual meta-reviews will remain unchanged, but the generated meta-reviews will be truncated.

\textbf{Experiment-4:} Both actual and generated will be truncated.
\end{displayquote}
\begin{table}
\centering
\caption{Results of ablation study 1 for meta-review generation}
\label{tab:meta_ablation1}
\begin{tabular}{|c|ccccc|}
\hline
Experiment &
  \multicolumn{5}{c|}{Experimental   Results} \\ \hline
\multirow{7}{*}{1} &
  \multicolumn{1}{c|}{\textbf{Model}} &
  \multicolumn{1}{c|}{\textbf{ROUGE 1}} &
  \multicolumn{1}{c|}{\textbf{ROUGE 2}} &
  \multicolumn{1}{c|}{\textbf{ROUGE L}} &
  \textbf{ROUGE L-sum} \\ \hhline{~-----} 
 &
  \multicolumn{1}{c|}{BERT} &
  \multicolumn{1}{c|}{0.337877671} &
  \multicolumn{1}{c|}{0.06955843} &
  \multicolumn{1}{c|}{0.187107362} &
  0.224058732 \\ \hhline{~-----} 
 &
  \multicolumn{1}{c|}{GPT2} &
  \multicolumn{1}{c|}{0.341020189} &
  \multicolumn{1}{c|}{0.071843498} &
  \multicolumn{1}{c|}{0.187573327} &
  0.225613879 \\ \hhline{~-----} 
 &
  \multicolumn{1}{c|}{BART} &
  \multicolumn{1}{c|}{0.331064783} &
  \multicolumn{1}{c|}{0.06898763} &
  \multicolumn{1}{c|}{0.192096169} &
  0.220325255 \\ \hhline{~-----} 
 &
  \multicolumn{1}{c|}{Pegasus} &
  \multicolumn{1}{c|}{0.163075545} &
  \multicolumn{1}{c|}{0.035384193} &
  \multicolumn{1}{c|}{0.116634477} &
  0.12564211 \\ \hhline{~-----} 
 &
  \multicolumn{1}{c|}{T5} &
  \multicolumn{1}{c|}{0.307273636} &
  \multicolumn{1}{c|}{0.06256972} &
  \multicolumn{1}{c|}{0.184502463} &
  0.208718978 \\ \hhline{~-----} 
 &
  \multicolumn{1}{c|}{T5 - Fine Tuned} &
  \multicolumn{1}{c|}{\textbf{0.358064672}} &
  \multicolumn{1}{c|}{\textbf{0.093864946}} &
  \multicolumn{1}{c|}{\textbf{0.222214337}} &
  \textbf{0.244906182} \\ \hline
\multirow{7}{*}{2} &
  \multicolumn{1}{c|}{\textbf{Model}} &
  \multicolumn{1}{c|}{\textbf{ROUGE 1}} &
  \multicolumn{1}{c|}{\textbf{ROUGE 2}} &
  \multicolumn{1}{c|}{\textbf{ROUGE L}} &
  \textbf{ROUGE L-sum} \\ \hhline{~-----} 
 &
  \multicolumn{1}{c|}{BERT} &
  \multicolumn{1}{c|}{0.289976204} &
  \multicolumn{1}{c|}{0.059633494} &
  \multicolumn{1}{c|}{0.157379733} &
  0.231071298 \\ \hhline{~-----} 
 &
  \multicolumn{1}{c|}{GPT2} &
  \multicolumn{1}{c|}{0.293075447} &
  \multicolumn{1}{c|}{0.061508612} &
  \multicolumn{1}{c|}{0.157923605} &
  0.233376565 \\ \hhline{~-----} 
 &
  \multicolumn{1}{c|}{BART} &
  \multicolumn{1}{c|}{0.2810028} &
  \multicolumn{1}{c|}{0.058560369} &
  \multicolumn{1}{c|}{0.160231548} &
  0.22236455 \\ \hhline{~-----} 
 &
  \multicolumn{1}{c|}{Pegasus} &
  \multicolumn{1}{c|}{0.130814274} &
  \multicolumn{1}{c|}{0.029305916} &
  \multicolumn{1}{c|}{0.092879966} &
  0.113197222 \\ \hhline{~-----} 
 &
  \multicolumn{1}{c|}{T5} &
  \multicolumn{1}{c|}{0.258477925} &
  \multicolumn{1}{c|}{0.052870843} &
  \multicolumn{1}{c|}{0.152796889} &
  0.205600646 \\ \hhline{~-----} 
 &
  \multicolumn{1}{c|}{T5 - Fine Tuned} &
  \multicolumn{1}{c|}{\textbf{0.304676699}} &
  \multicolumn{1}{c|}{\textbf{0.077178461}} &
  \multicolumn{1}{c|}{\textbf{0.185579184}} &
  \textbf{0.237612471} \\ \hline
\multirow{7}{*}{3} &
  \multicolumn{1}{c|}{\textbf{Model}} &
  \multicolumn{1}{c|}{\textbf{ROUGE 1}} &
  \multicolumn{1}{c|}{\textbf{ROUGE 2}} &
  \multicolumn{1}{c|}{\textbf{ROUGE L}} &
  \textbf{ROUGE L-sum} \\ \hhline{~-----} 
 &
  \multicolumn{1}{c|}{BERT} &
  \multicolumn{1}{c|}{0.331008893} &
  \multicolumn{1}{c|}{0.068244777} &
  \multicolumn{1}{c|}{\textbf{0.183123335}} &
  0.251485305 \\ \hhline{~-----} 
 &
  \multicolumn{1}{c|}{GPT2} &
  \multicolumn{1}{c|}{\textbf{0.332831242}} &
  \multicolumn{1}{c|}{0.070232848} &
  \multicolumn{1}{c|}{0.182957254} &
  \textbf{0.252544428} \\ \hhline{~-----} 
 &
  \multicolumn{1}{c|}{BART} &
  \multicolumn{1}{c|}{0.310536194} &
  \multicolumn{1}{c|}{0.064757} &
  \multicolumn{1}{c|}{0.180140606} &
  0.237876746 \\ \hhline{~-----} 
 &
  \multicolumn{1}{c|}{Pegasus} &
  \multicolumn{1}{c|}{0.153857284} &
  \multicolumn{1}{c|}{0.03383729} &
  \multicolumn{1}{c|}{0.110490665} &
  0.118747819 \\ \hhline{~-----} 
 &
  \multicolumn{1}{c|}{T5} &
  \multicolumn{1}{c|}{0.294395993} &
  \multicolumn{1}{c|}{0.060214324} &
  \multicolumn{1}{c|}{0.176710208} &
  0.229207545 \\ \hhline{~-----} 
 &
  \multicolumn{1}{c|}{T5 - Fine Tuned} &
  \multicolumn{1}{c|}{0.291653073} &
  \multicolumn{1}{c|}{\textbf{0.076178006}} &
  \multicolumn{1}{c|}{0.178689179} &
  0.231285738 \\ \hline
\multirow{7}{*}{4} &
  \multicolumn{1}{c|}{\textbf{Model}} &
  \multicolumn{1}{c|}{\textbf{ROUGE 1}} &
  \multicolumn{1}{c|}{\textbf{ROUGE 2}} &
  \multicolumn{1}{c|}{\textbf{ROUGE L}} &
  \textbf{ROUGE L-sum} \\ \hhline{~-----} 
 &
  \multicolumn{1}{c|}{BERT} &
  \multicolumn{1}{c|}{0.284088786} &
  \multicolumn{1}{c|}{0.058499394} &
  \multicolumn{1}{c|}{0.154010156} &
  0.250611715 \\ \hhline{~-----} 
 &
  \multicolumn{1}{c|}{GPT2} &
  \multicolumn{1}{c|}{\textbf{0.285807989}} &
  \multicolumn{1}{c|}{0.060096721} &
  \multicolumn{1}{c|}{\textbf{0.153866771}} &
  \textbf{0.252302857} \\ \hhline{~-----} 
 &
  \multicolumn{1}{c|}{BART} &
  \multicolumn{1}{c|}{0.264091179} &
  \multicolumn{1}{c|}{0.055059944} &
  \multicolumn{1}{c|}{0.150529308} &
  0.234939516 \\ \hhline{~-----} 
 &
  \multicolumn{1}{c|}{Pegasus} &
  \multicolumn{1}{c|}{0.12299954} &
  \multicolumn{1}{c|}{0.027932944} &
  \multicolumn{1}{c|}{0.087733046} &
  0.106466784 \\ \hhline{~-----} 
 &
  \multicolumn{1}{c|}{T5} &
  \multicolumn{1}{c|}{0.247783672} &
  \multicolumn{1}{c|}{0.050878349} &
  \multicolumn{1}{c|}{0.146427063} &
  0.222847712 \\ \hhline{~-----} 
 &
  \multicolumn{1}{c|}{T5 - Fine Tuned} &
  \multicolumn{1}{c|}{0.250180216} &
  \multicolumn{1}{c|}{\textbf{0.063146913}} &
  \multicolumn{1}{c|}{0.150333555} &
  0.223826561 \\ \hline
\end{tabular}%
\end{table}

\begin{table}
\centering
\caption{Results of ablation study 2 for meta-review generation}
\label{tab:meta_ablation2}
\begin{tabular}{|c|ccccc|}
\hline
Experiment &
  \multicolumn{5}{c|}{Experimental   Results} \\ \hline
\multirow{7}{*}{1} &
  \multicolumn{1}{c|}{\textbf{Model}} &
  \multicolumn{1}{c|}{\textbf{ROUGE 1}} &
  \multicolumn{1}{c|}{\textbf{ROUGE 2}} &
  \multicolumn{1}{c|}{\textbf{ROUGE L}} &
  \textbf{ROUGE L-sum} \\ \hhline{~-----} 
 &
  \multicolumn{1}{c|}{BERT} &
  \multicolumn{1}{c|}{0.309887606} &
  \multicolumn{1}{c|}{0.054499715} &
  \multicolumn{1}{c|}{0.158203052} &
  0.158203052 \\ \hhline{~-----} 
 &
  \multicolumn{1}{c|}{GPT2} &
  \multicolumn{1}{c|}{0.310100587} &
  \multicolumn{1}{c|}{0.056759171} &
  \multicolumn{1}{c|}{0.157663803} &
  0.157663803 \\ \hhline{~-----} 
 &
  \multicolumn{1}{c|}{BART} &
  \multicolumn{1}{c|}{0.312600646} &
  \multicolumn{1}{c|}{0.057880428} &
  \multicolumn{1}{c|}{0.177837056} &
  0.177837056 \\ \hhline{~-----} 
 &
  \multicolumn{1}{c|}{Pegasus} &
  \multicolumn{1}{c|}{0.129868659} &
  \multicolumn{1}{c|}{0.023221815} &
  \multicolumn{1}{c|}{0.09481926} &
  0.09481926 \\ \hhline{~-----} 
 &
  \multicolumn{1}{c|}{T5} &
  \multicolumn{1}{c|}{0.273136181} &
  \multicolumn{1}{c|}{0.046187633} &
  \multicolumn{1}{c|}{0.164578858} &
  0.164578858 \\ \hhline{~-----} 
 &
  \multicolumn{1}{c|}{T5 - Fine Tuned} &
  \multicolumn{1}{c|}{\textbf{0.319566243}} &
  \multicolumn{1}{c|}{\textbf{0.075371796}} &
  \multicolumn{1}{c|}{\textbf{0.203016246}} &
  \textbf{0.203016246} \\ \hline
\multirow{7}{*}{2} &
  \multicolumn{1}{c|}{\textbf{Model}} &
  \multicolumn{1}{c|}{\textbf{ROUGE 1}} &
  \multicolumn{1}{c|}{\textbf{ROUGE 2}} &
  \multicolumn{1}{c|}{\textbf{ROUGE L}} &
  \textbf{ROUGE L-sum} \\ \hhline{~-----} 
 &
  \multicolumn{1}{c|}{BERT} &
  \multicolumn{1}{c|}{0.269063644} &
  \multicolumn{1}{c|}{0.047170521} &
  \multicolumn{1}{c|}{0.134643407} &
  0.134643407 \\ \hhline{~-----} 
 &
  \multicolumn{1}{c|}{GPT2} &
  \multicolumn{1}{c|}{\textbf{0.269738353}} &
  \multicolumn{1}{c|}{0.049275396} &
  \multicolumn{1}{c|}{0.134404076} &
  0.134404076 \\ \hhline{~-----} 
 &
  \multicolumn{1}{c|}{BART} &
  \multicolumn{1}{c|}{0.264809121} &
  \multicolumn{1}{c|}{0.049236616} &
  \multicolumn{1}{c|}{0.148078968} &
  0.148078968 \\ \hhline{~-----} 
 &
  \multicolumn{1}{c|}{Pegasus} &
  \multicolumn{1}{c|}{0.102774924} &
  \multicolumn{1}{c|}{0.018773986} &
  \multicolumn{1}{c|}{0.074461183} &
  0.074461183 \\ \hhline{~-----} 
 &
  \multicolumn{1}{c|}{T5} &
  \multicolumn{1}{c|}{0.230471272} &
  \multicolumn{1}{c|}{0.039238822} &
  \multicolumn{1}{c|}{0.13681206} &
  0.13681206 \\ \hhline{~-----} 
 &
  \multicolumn{1}{c|}{T5 - Fine Tuned} &
  \multicolumn{1}{c|}{0.268759407} &
  \multicolumn{1}{c|}{\textbf{0.061845592}} &
  \multicolumn{1}{c|}{\textbf{0.168703117}} &
  \textbf{0.168703117} \\ \hline
\multirow{7}{*}{3} &
  \multicolumn{1}{c|}{\textbf{Model}} &
  \multicolumn{1}{c|}{\textbf{ROUGE 1}} &
  \multicolumn{1}{c|}{\textbf{ROUGE 2}} &
  \multicolumn{1}{c|}{\textbf{ROUGE L}} &
  \textbf{ROUGE L-sum} \\ \hhline{~-----} 
 &
  \multicolumn{1}{c|}{BERT} &
  \multicolumn{1}{c|}{0.275143692} &
  \multicolumn{1}{c|}{0.048762729} &
  \multicolumn{1}{c|}{0.138440669} &
  0.138440669 \\ \hhline{~-----} 
 &
  \multicolumn{1}{c|}{GPT2} &
  \multicolumn{1}{c|}{0.27557688} &
  \multicolumn{1}{c|}{0.050477718} &
  \multicolumn{1}{c|}{0.137889717} &
  0.137889717 \\ \hhline{~-----} 
 &
  \multicolumn{1}{c|}{BART} &
  \multicolumn{1}{c|}{\textbf{0.286304383}} &
  \multicolumn{1}{c|}{\textbf{0.053320322}} &
  \multicolumn{1}{c|}{\textbf{0.162855943}} &
  \textbf{0.162855943} \\ \hhline{~-----} 
 &
  \multicolumn{1}{c|}{Pegasus} &
  \multicolumn{1}{c|}{0.129410031} &
  \multicolumn{1}{c|}{0.02316823} &
  \multicolumn{1}{c|}{0.094485449} &
  0.094485449 \\ \hhline{~-----} 
 &
  \multicolumn{1}{c|}{T5} &
  \multicolumn{1}{c|}{0.244862433} &
  \multicolumn{1}{c|}{0.041870821} &
  \multicolumn{1}{c|}{0.147229548} &
  0.147229548 \\ \hhline{~-----} 
 &
  \multicolumn{1}{c|}{T5 - Fine Tuned} &
  \multicolumn{1}{c|}{0.186406627} &
  \multicolumn{1}{c|}{0.043953507} &
  \multicolumn{1}{c|}{0.11616212} &
  0.11616212 \\ \hline
\multirow{7}{*}{4} &
  \multicolumn{1}{c|}{\textbf{Model}} &
  \multicolumn{1}{c|}{\textbf{ROUGE 1}} &
  \multicolumn{1}{c|}{\textbf{ROUGE 2}} &
  \multicolumn{1}{c|}{\textbf{ROUGE L}} &
  \textbf{ROUGE L-sum} \\ \hhline{~-----} 
 &
  \multicolumn{1}{c|}{BERT} &
  \multicolumn{1}{c|}{0.240032203} &
  \multicolumn{1}{c|}{0.042386221} &
  \multicolumn{1}{c|}{0.118324455} &
  0.118324455 \\ \hhline{~-----} 
 &
  \multicolumn{1}{c|}{GPT2} &
  \multicolumn{1}{c|}{0.240868716} &
  \multicolumn{1}{c|}{0.044032717} &
  \multicolumn{1}{c|}{0.118075546} &
  0.118075546 \\ \hhline{~-----} 
 &
  \multicolumn{1}{c|}{BART} &
  \multicolumn{1}{c|}{\textbf{0.24338546}} &
  \multicolumn{1}{c|}{\textbf{0.045600195}} &
  \multicolumn{1}{c|}{\textbf{0.136143933}} &
  \textbf{0.136143933} \\ \hhline{~-----} 
 &
  \multicolumn{1}{c|}{Pegasus} &
  \multicolumn{1}{c|}{0.102383776} &
  \multicolumn{1}{c|}{0.018725795} &
  \multicolumn{1}{c|}{0.074183734} &
  0.074183734 \\ \hhline{~-----} 
 &
  \multicolumn{1}{c|}{T5} &
  \multicolumn{1}{c|}{0.206212947} &
  \multicolumn{1}{c|}{0.035512903} &
  \multicolumn{1}{c|}{0.122157236} &
  0.122157236 \\ \hhline{~-----} 
 &
  \multicolumn{1}{c|}{T5 - Fine Tuned} &
  \multicolumn{1}{c|}{0.158161315} &
  \multicolumn{1}{c|}{0.036207722} &
  \multicolumn{1}{c|}{0.097111502} &
  0.097111502 \\ \hline
\end{tabular}%
\end{table}

\noindent
In this study, the first two experiments are of particular significance. Experiment 1 serves as a benchmark for the performance of the model by comparing its output to a standard reference in the form of an actual meta-review. On the other hand, Experiment 2 aims to assess the ability of the model to generalize the knowledge it has acquired during the training phase by evaluating the degree of similarity between its outputs and the training data. These experiments provide valuable insights into the capabilities of the model and are essential in determining its performance for practical applications. Experiment 2 is needed to make the comparison a valid one as we fine-tune the model using a truncated meta-review. Despite this, the last two experiments were carried out to observe their performance to complete the tests. The results of the ablation studies are given in Table \ref{tab:meta_ablation1} and \ref{tab:meta_ablation2}.\\
\noindent
The results of the ablation studies are presented in \ref{tab:meta_ablation1} and \ref{tab:meta_ablation2}. A comprehensive analysis of the experimental results shows that the fine-tuned T5 model outperforms the inference model in experiments 1 and 2 in both ablation studies. This superiority is observed across all metrics used to evaluate the generated meta-reviews. Additionally, the results indicate that the GPT2 model generated scores that are highly competitive with those of the fine-tuned T5 model across all metrics, while the performance of the Pegasus model was found to be the lowest among the models. Though it is clear from the experimental results that fine-tuned T5 is significantly better compared to the inference models in experiments 1, and 2, we perform a statistical test Wilcoxon signed-rank test \cite{woolson2007wilcoxon} based on all the metrics to validate the findings. From the statistical test, we validate our findings from the experimental results.\\
\noindent
The observation that the results of ablation study 1 are superior to those of ablation study 2 is of particular significance. In ablation study 1, the TextRank approach was used for truncating the peer reviews during fine-tuning. Our experiments have shown that TextRank provides more effective summarization than TextRank with sentiment analysis, which explains the superior results obtained in ablation study 1. In addition, we get the best results when we truncate neither generated nor reference meta-review. This represents that we do not need to truncate the generated meta-review to get the optimal result.  \\
\noindent
 Further research may be necessary to validate and expand upon these findings. Fine-tuning the ratios of the positive, negative, and neutral sentences in the TextRank with sentiment analysis may yield better performance of the model. Furthermore, the sentiment analysis could not pick up on the positive and negative sentences properly due to the way a peer review is written. So, fine-tuning it for the dataset may yield better performance as well.

\subsubsection{Comparison with existing methods}
\begin{table}[]
\centering
\caption{Performance comparison between MRGen (with attention) \cite{pradhan2021deep} and our proposed method of acceptance decision prediction}
\label{tab:task1_comparison}
\begin{tabular}{|c|l|c|}
\hline
\textbf{Dataset}           & \multicolumn{1}{c|}{\textbf{Models}} & \textbf{Accuracy} \\ \hline
\multirow{2}{*}{ICLR-2017} & BERT + Decision Tree                 & 86.80\%           \\ \hhline{~--} 
                           & MRGen (with attention)               & 81.50\%           \\ \hline
\multirow{2}{*}{ICLR-2018} & BERT + Decision Tree                 & 87.10\%           \\ \hhline{~--} 
                           & MRGen (with attention)               & 84.60\%           \\ \hline
\multirow{2}{*}{ICLR-2019} & BERT + Decision Tree                 & 88.90\%           \\ \hhline{~--} 
                           & MRGen (with attention)               & 85.80\%           \\ \hline
\end{tabular}
\end{table}

For the acceptance decision prediction, the current state-of-the-art model is MRGen with attention  \cite{pradhan2021deep}. The best accuracy score of our experimental analysis outperforms all of the existing models for acceptance decision prediction. The maximum accuracy of MRGen with attention for this task is 85.8\%. But we achieve an 87.62\% accuracy score for the PeerRead \cite{kang2018dataset} dataset using Decision Tree as the classifier and BERT as the word embedding technique. In addition, we get an accuracy score of 88.02\% on the merged dataset. Authors of MRGen with attention use ICLR 2017 -– 2019 separately, but we use a merged dataset for all the experiments. That is why to make a valid comparison, we utilize BERT and Decision Tree combination in ICLR 2017, 2018, and 2019 separately for training and testing as per the research work of T. Pradhan et al. \cite{pradhan2021deep}. The results are shown in Table \ref{tab:task1_comparison}. We use feature set 7 (paper title, peer review, recommendation score, and confidence score) for these experiments as we get the best results in terms of accuracy and precision using this feature set. The experimental results \ref{tab:task1_comparison} show that the combination of BERT as the word embedding technique and Decision Tree as the classifier outperforms the existing state-of-the-art results on ICLR 2017, 2018, and 2019. We get the accuracy scores of 86.80\%, 87.10\%, and 88.90\% on ICLR 2017, 2018, and 2019 where the previous best accuracy scores obtained by MRGen (with attention) were in order 81.50\%, 84.60\%, and 85.80\%. Wilcoxon signed-rank test \cite{woolson2007wilcoxon} shows significant improvement in our result compared with MRGen (with attention) \cite{pradhan2021deep}. \\
In addition, we use the dataset of other international conferences and traditional machine learning algorithms that require less time for the training phase and can generate the decision faster. Moreover, we report the F1 score along with the accuracy score as it is one of the most important evaluation matrices for the classification tasks which are based on precision and recall scores.\\
\begin{table}[]
\centering
\caption{Performance comparison between MRGen \cite{pradhan2021deep} and our proposed method of meta-review generation}
\label{tab:task2_comparison}
\begin{tabular}{|c|c|c|c|c|c|}
\hline
\textbf{Dataset} & \textbf{Method} & \textbf{ROUGE 1} & \textbf{ROUGE 2} & \textbf{ROUGE L} & \textbf{Readability} \\ \hline
\multirow{2}{*}{ICLR-2017} & T5 - Fine Tuned & 38.95 & 13.79 & 26.62 & 15.21 \\ \hhline{~-----} 
                           & MRGen           & 43.55 & 19.68 & 30.98 & 14.76 \\ \hline
\multirow{2}{*}{ICLR-2018} & T5 - Fine Tuned & 35.75 & 9.41  & 21.67 & 15.29 \\ \hhline{~-----} 
                           & MRGen           & 31.31 & 5.78  & 21.20 & 14.59 \\ \hline
\multirow{2}{*}{ICLR-2019} & T5 - Fine Tuned & 36.16 & 9.09  & 23.72 & 15.33 \\ \hhline{~-----} 
                           & MRGen           & 34.84 & 7.33  & 22.95 & 14.84 \\ \hline
\end{tabular}
\end{table}
\noindent
For the meta-review generation, one of the most relevant model MRGen \cite{pradhan2021deep} for the meta-review generation gets good scores on specific cases compared to us as we can see from the experimental results in Table \ref{tab:task2_comparison}. In Table \ref{tab:task2_comparison} MRGen \cite{pradhan2021deep} is compared with the T5-Fine Tuned method where we use TextRank algorithm to truncate the Peer Reviews and the meta-review at the time of training and no truncation of reviews on the time of comparison. Experimental results show that the T5- Fine Tuned method surpasses the results of MRGen in terms of ROUGE scores on ICLR 2018 and 2019 datasets. Moreover, we can see from the results that T5 - Fine Tuned generated meta-reviews are more readable than MRGen-generated meta-reviews. To evaluate the meta-review in terms of readability, we use the Coleman–Liau index from \textit{ReadabilityCalculator 0.2.37} from PyPI as them \cite{pradhan2021deep}. Wilcoxon signed-rank test \cite{woolson2007wilcoxon} shows significant improvement in our result in terms of readability compared with MRGen \cite{pradhan2021deep}. In addition, we use a large dataset which is a combination of ICLR 2017 -- 2021, CORL 2021, and GI 2020 in our experimental analysis to make a uniform model for the meta-review generation.\\
\noindent
We use a large and merged dataset using information from the different international conferences for the completion of the peer review aggregation task. The accuracy score we get surpasses the existing methods in terms of acceptance decision prediction. Moreover, for the meta-review generation, we use the transfer learning approach and get a very competitive score.

\section{Conclusion and Future Work}
The number of papers submitted to top-tier conferences is constantly increasing. It is becoming more difficult for meta-reviewers to deliver a compatible meta-review and acceptance judgment. After reviewing the previous works it is clear that there are very few works available and there is no clear-cut solution for the meta-review generation process. To vectorize the sentences for the word embedding, we used BERT as BERT is currently the state-of-the-art for natural language understanding tasks. For experimental analysis, we used Word2Vec and fastText as the word embedding techniques. For the acceptance decision prediction, we used traditional machine learning algorithms KNN, Decision Tree, Random Forest, Logistic Regression, Naive Bayes, and SVM Linear. We got the best accuracy and F1 score using Random Forest and BERT as word embedding techniques on the whole dataset. Moreover, we did an ablation study, and from that, we figured out that the paper title and abstract are not that important for acceptance decision prediction. We need to be very careful while considering confidence score as the feature because it might create noise while predicting the decision. The recommendation score plays a vital role in this task. For the meta-review generation task, we have fine-tuned Google's T5 Transformer on our dataset for our work to generate the abstractive summary of the peer reviews. In addition, we used inference models BERT, GPT2, BART, Pegasus, and T5 for the experimental analysis. To avoid the resource allocation problem because of the length of peer reviews and meta-reviews, we used the \textit{TextRank} algorithm to truncate the peer reviews. In addition, to truncate the peer reviews we used \textit{TextRank} algorithm with sentiment analysis as well. For the ablation study, we used truncation of meta-reviews while evaluating. We figured out that fine-tuned T5 performs better than inference models, in addition, only \textit{TextRank} algorithm gives the effective truncation compared to other techniques.\\
\noindent
In this study, we used traditional machine learning algorithms for acceptance decision prediction. As an extension of our work, deep learning techniques i.e. Bi-LSTM, Bi-LSTM with an attention mechanism can be used to compare with the results of traditional machine learning algorithms. As we were getting satisfactory results using the traditional machine learning algorithms, we did not use any complex architecture for the acceptance decision prediction as it requires more time for training. For the meta-review generation, other transformer models can be fine-tuned i.e. BART, GPT2 as we were getting close to the best results with the inference models of these models. In addition, other mechanisms can be explored for truncating reviews to increase efficiency.  In recent times, there have been a few advancements in assessing a peer review's quality \cite{mah2023art} \cite{meng2023assessing}. Incorporating a method to utilize higher-quality reviews more effectively for meta-review generation and acceptance decision prediction is a research topic that can be explored to yield high-quality meta-reviews.\\


{
\small

}


\newpage
\appendix

\section{Appendix}

\subsection{Ablation Study Results of Acceptance Decision Prediction}
\begin{table}[!htbp]
\centering
\caption{Results of Ablation Study 2 for acceptance decision prediction using PeerRead \cite{kang2018dataset} for training and Conference [1, 5] for testing}
\label{tab:experiment 2}
\scriptsize
\begin{tabular}{|c|c|*{6}{c|}}
\hline
{Exp.} & \multicolumn{7}{c|}{Experimental Results} \\ \hline

\multirow{9}{*}{1} & \multirow{3}{*}{Algorithm} & \multicolumn{6}{c|}{Word Embedding Technique} \\ \hhline{~~------} 
 &  & \multicolumn{2}{c|}{BERT} & \multicolumn{2}{c|}{Word2Vec} & \multicolumn{2}{c|}{fastText} \\ \hhline{~~------} 
 &  & \textbf{ACC (\%)} & \textbf{F1 (\%)} & \textbf{ACC (\%)} & \textbf{F1 (\%)} & \textbf{ACC (\%)} & \textbf{F1 (\%)} \\ \hhline{~-------} 
 & KNN (K = 5) & 59.88 & 27.39 & 58.58 & 56.68 & \textbf{69.57} & \textbf{58.54} \\ \hhline{~-------} 
 & Decision Tree & 45.89 & 45.36 & 53.05 & 49.78 & 47.03 & 45.52 \\ \hhline{~-------} 
 & Random Forest & 54.11 & 31.55 & 48.17 & 32.88 & 52.73 & 34.2 \\ \hhline{~-------} 
 & Logistic Regression & 57.53 & 54.53 & 60.62 & 56.79 & 61.27 & 57.27 \\ \hhline{~-------} 
 & Naive Bayes & 54.27 & 32.61 & 63.14 & 52.66 & 56.39 & 41.23 \\ \hhline{~-------} 
 & SVM Linear & 57.85 & 54.72 & 60.13 & 56.41 & 58.1 & 55.49 \\ \hline

\multirow{9}{*}{2} & \multirow{3}{*}{Algorithm} & \multicolumn{6}{c|}{Word Embedding Technique} \\ \hhline{~~------} 
 &  & \multicolumn{2}{c|}{BERT} & \multicolumn{2}{c|}{Word2Vec} & \multicolumn{2}{c|}{fastText} \\ \hhline{~~------} 
 &  & \textbf{ACC (\%)} & \textbf{F1 (\%)} & \textbf{ACC (\%)} & \textbf{F1 (\%)} & \textbf{ACC (\%)} & \textbf{F1 (\%)} \\ \hhline{~-------} 
 & KNN (K = 5) & 64.28 & 29.08 & 87.63 & 79.46 & 78.11 & 59.3 \\ \hhline{~-------} 
 & Decision Tree & 61.43 & 52.41 & 64.85 & 57.89 & 66.15 & 57.46 \\ \hhline{~-------} 
 & Random Forest & 62.9 & 31.33 & 47.93 & 31.48 & 59.8 & 27.99 \\ \hhline{~-------} 
 & Logistic Regression & 85.76 & 74.74 & 84.7 & 70.53 & 81.94 & 63.37 \\ \hhline{~-------} 
 & Naive Bayes & 54.68 & 32.81 & 63.3 & 52.38 & 56.71 & 41.15 \\ \hhline{~-------} 
 & SVM Linear & \textbf{89.83} & \textbf{81.8} & 86.82 & 75.15 & 85.35 & 73.05 \\ \hline

\multirow{9}{*}{3} & \multirow{3}{*}{Algorithm} & \multicolumn{6}{c|}{Word Embedding Technique} \\ \hhline{~~------} 
 & & \multicolumn{2}{c|}{BERT} & \multicolumn{2}{c|}{Word2Vec} & \multicolumn{2}{c|}{fastText} \\ \hhline{~~------} 
 & & \textbf{ACC (\%)} & \textbf{F1 (\%)} & \textbf{ACC (\%)} & \textbf{F1 (\%)} & \textbf{ACC (\%)} & \textbf{F1 (\%)} \\ \hhline{~-------} 
 & KNN (K = 5) & 59.48 & 20.95 & 65.26 & 22.78 & 62.08 & 23.61 \\ \hhline{~-------} 
 & Decision Tree & 59.24 & 19.06 & 48.58 & 29.31 & 57.36 & 33.33 \\ \hhline{~-------} 
 & Random Forest & 60.62 & 12.63 & 53.13 & 24.01 & 45.57 & 21.02 \\ \hhline{~-------} 
 & Logistic Regression & 65.01 & 5.29 & \textbf{72.01} & 26.81 & 71.68 & 27.5 \\ \hhline{~-------} 
 & Naive Bayes & 54.6 & 32.45 & 63.22 & \textbf{51.5} & 56.96 & 40.09 \\ \hhline{~-------} 
 & SVM Linear & 67.21 & 4.28 & 69.81 & 4.63 & 71.36 & 19.63 \\ \hline

\multirow{9}{*}{4} & \multirow{3}{*}{Algorithm} & \multicolumn{6}{c|}{Word Embedding Technique} \\ \hhline{~~------} 
 & & \multicolumn{2}{c|}{BERT} & \multicolumn{2}{c|}{Word2Vec} & \multicolumn{2}{c|}{fastText} \\ \hhline{~~------} 
 & & \textbf{ACC (\%)} & \textbf{F1 (\%)} & \textbf{ACC (\%)} & \textbf{F1 (\%)} & \textbf{ACC (\%)} & \textbf{F1 (\%)} \\ \hhline{~-------} 
 & KNN (K = 5) & 54.84 & 49.86 & 57.93 & 54.85 & \textbf{67.86} & \textbf{58.02} \\ \hhline{~-------} 
 & Decision Tree & 51.34 & 49.06 & 51.02 & 46.35 & 48.41 & 44.48 \\ \hhline{~-------} 
 & Random Forest & 50.69 & 49.16 & 51.83 & 39.34 & 58.67 & 42.27 \\ \hhline{~-------} 
 & Logistic Regression & 58.01 & 54.97 & 59.97 & 56.31 & 61.84 & 57.71 \\ \hhline{~-------} 
 & Naive Bayes & 44.59 & 41.75 & 62.41 & 49.89 & 53.46 & 52.17 \\ \hhline{~-------} 
 & SVM Linear & 58.01 & 54.82 & 60.86 & 56.94 & 60.21 & 56.61 \\ \hline

\multirow{9}{*}{5} & \multirow{3}{*}{Algorithm} & \multicolumn{6}{c|}{Word Embedding Technique} \\ \hhline{~~------} 
 & & \multicolumn{2}{c|}{BERT} & \multicolumn{2}{c|}{Word2Vec} & \multicolumn{2}{c|}{fastText} \\ \hhline{~~------} 
 & & \textbf{ACC (\%)} & \textbf{F1 (\%)} & \textbf{ACC (\%)} & \textbf{F1 (\%)} & \textbf{ACC (\%)} & \textbf{F1 (\%)} \\ \hhline{~-------} 
 & KNN (K = 5) & 83.24 & 73.04 & 88.2 & 80.48 & 74.69 & 59.13 \\ \hhline{~-------} 
 & Decision Tree & 78.11 & 67.63 & 72.34 & 61.97 & 67.7 & 59.03 \\ \hhline{~-------} 
 & Random Forest & 72.66 & 59.12 & 49.15 & 32.58 & 59.72 & 31.35 \\ \hhline{~-------} 
 & Logistic Regression & 88.52 & 79.36 & 85.27 & 71.67 & 81.69 & 62.31 \\ \hhline{~-------} 
 & Naive Bayes & 44.75 & 41.81 & 63.06 & 49.44 & 56.22 & 51.96 \\ \hhline{~-------} 
 & SVM Linear & \textbf{89.5} & \textbf{80.66} & 86.33 & 73.99 & 84.95 & 70.59 \\ \hline

\multirow{9}{*}{6} & \multirow{3}{*}{Algorithm} & \multicolumn{6}{c|}{Word Embedding Technique} \\ \hhline{~~------} 
 &  & \multicolumn{2}{c|}{BERT} & \multicolumn{2}{c|}{Word2Vec} & \multicolumn{2}{c|}{fastText} \\ \hhline{~~------} 
 &  & \textbf{ACC (\%)} & \textbf{F1 (\%)} & \textbf{ACC (\%)} & \textbf{F1 (\%)} & \textbf{ACC (\%)} & \textbf{F1 (\%)} \\ \hhline{~-------} 
 & KNN (K = 5) & 65.91 & 22.26 & 68.84 & 21.36 & 58.42 & 15.54 \\ \hhline{~-------} 
 & Decision Tree & 66.96 & 4.69 & 56.88 & 23.85 & 59.07 & 30.43 \\ \hhline{~-------} 
 & Random Forest & 67.05 & 5.59 & 46.95 & 28.98 & 64.93 & 18.83 \\ \hhline{~-------} 
 & Logistic Regression & 67.29 & 6.07 & \textbf{72.01} & 28.33 & 71.28 & 25.68 \\ \hhline{~-------} 
 & Naive Bayes & 44.83 & 41.85 & 63.14 & \textbf{48.58} & 54.35 & 47.91 \\ \hhline{~-------} 
 & SVM Linear & 67.21 & 4.28 & 69.16 & 0.53 & 69.98 & 6.58 \\ \hline

\multirow{9}{*}{7} & \multirow{3}{*}{Algorithm} & \multicolumn{6}{c|}{Word Embedding Technique} \\ \hhline{~~------} 
 &  & \multicolumn{2}{c|}{BERT} & \multicolumn{2}{c|}{Word2Vec} & \multicolumn{2}{c|}{fastText} \\ \hhline{~~------} 
 &  & \textbf{ACC (\%)} & \textbf{F1 (\%)} & \textbf{ACC (\%)} & \textbf{F1 (\%)} & \textbf{ACC (\%)} & \textbf{F1 (\%)} \\ \hhline{~-------} 
 & KNN (K = 5) & 58.83 & 43.53 & 57.36 & 54.36 & \textbf{67.29} & 57.32 \\ \hhline{~-------} 
 & Decision Tree & 45.4 & 41.6 & 42.23 & 44.01 & 50.12 & 44.22 \\ \hhline{~-------} 
 & Random Forest & 53.21 & 44.01 & 50.85 & 35.33 & 66.88 & 41.94 \\ \hhline{~-------} 
 & Logistic Regression & 57.53 & 54.69 & 60.13 & 56.48 & 62 & \textbf{57.81} \\ \hhline{~-------} 
 & Naive Bayes & 52.24 & 36.54 & 62.65 & 51.12 & 50.85 & 45.09 \\ \hhline{~-------} 
 & SVM Linear & 57.69 & 54.86 & 60.78 & 56.89 & 60.21 & 56.61 \\ \hline

\end{tabular}
\end{table}

\begin{table}[htbp]
\centering
\caption{Results of Ablation Study 3 for acceptance decision prediction using a merged dataset of PeerRead \cite{kang2018dataset} and Conference [1, 5]}
\label{tab:experiment 3}
\scriptsize
\begin{tabular}{|c|c|*{6}{c|}}
\hline
{Exp.} & \multicolumn{7}{c|}{Experimental Results} \\ \hline

\multirow{9}{*}{1} & {\multirow{3}{*}{Algorithm}} & \multicolumn{6}{c|}{Word Embedding Technique} \\ \hhline{~~------} 
 & & \multicolumn{2}{c|}{BERT} & \multicolumn{2}{c|}{Word2Vec} & \multicolumn{2}{c|}{fastText} \\ \hhline{~~------} 
 & & \textbf{ACC (\%)} & \textbf{F1 (\%)} & \textbf{ACC (\%)} & \textbf{F1 (\%)} & \textbf{ACC (\%)} & \textbf{F1 (\%)} \\ \hhline{~-------} 
 & KNN (K = 5) & 66.26 & 34.29 & \textbf{86.8} & \textbf{79.55} & 77.51 & 60.34 \\ \hhline{~-------} 
 & Decision Tree & 85.21 & 78.2 & 81.78 & 72.86 & 80.56 & 71.35 \\ \hhline{~-------} 
 & Random Forest & 71.03 & 45.27 & 71.88 & 36.11 & 74.94 & 46.48 \\ \hhline{~-------} 
 & Logistic Regression & 81.05 & 68.3 & 84.47 & 74.75 & 82.64 & 71.71 \\ \hhline{~-------} 
 & Naive Bayes & 51.1 & 45.5 & 52.44 & 50.44 & 49.76 & 47.91 \\ \hhline{~-------} 
 & SVM Linear & 81.54 & 67.53 & 84.11 & 74.51 & 83.37 & 73.33 \\ \hline

\multirow{9}{*}{2} & \multirow{3}{*}{Algorithm} & \multicolumn{6}{c|}{Word Embedding Technique} \\ \hhline{~~------} 
 & & \multicolumn{2}{c|}{BERT} & \multicolumn{2}{c|}{Word2Vec} & \multicolumn{2}{c|}{fastText} \\ \hhline{~~------} 
 & & \textbf{ACC (\%)} & \textbf{F1 (\%)} & \textbf{ACC (\%)} & \textbf{F1 (\%)} & \textbf{ACC (\%)} & \textbf{F1 (\%)} \\ \hhline{~-------} 
 & KNN (K = 5) & 64.91 & 32.47 & 78.36 & 64.24 & 71.39 & 49.57 \\ \hhline{~-------} 
 & Decision Tree & 74.82 & 62.82 & 74.33 & 62.09 & 71.27 & 57.35 \\ \hhline{~-------} 
 & Random Forest & 67.35 & 42.33 & 70.78 & 32.68 & 71.52 & 37.53 \\ \hhline{~-------} 
 & Logistic   Regression & 80.8 & 67.22 & 78.73 & 64.5 & 78.85 & 64.18 \\ \hhline{~-------} 
 & Naive Bayes & 50.98 & 44.23 & 52.44 & 50.19 & 49.39 & 47.06 \\ \hhline{~-------} 
 & SVM Linear & \textbf{81.66} & \textbf{67.95} & 80.92 & 66.95 & 80.07 & 65.54 \\ \hline

\multirow{9}{*}{3} & \multirow{3}{*}{Algorithm} & \multicolumn{6}{c|}{Word Embedding Technique} \\ \hhline{~~------} 
 &  & \multicolumn{2}{c|}{BERT} & \multicolumn{2}{c|}{Word2Vec} & \multicolumn{2}{c|}{fastText} \\ \hhline{~~------} 
 &  & \textbf{ACC (\%)} & \textbf{F1 (\%)} & \textbf{ACC (\%)} & \textbf{F1 (\%)} & \textbf{ACC (\%)} & \textbf{F1 (\%)} \\ \hhline{~-------} 
 & KNN (K = 5) & 59.41 & 22.07 & 61.6 & 33.19 & 62.47 & 32.82 \\ \hhline{~-------} 
 & Decision Tree & 60.39 & 20.59 & 61.25 & 39.62 & 62.22 & 42.67 \\ \hhline{~-------} 
 & Random Forest & 62.35 & 22.61 & 68.58 & 26.78 & 69.19 & 27.59 \\ \hhline{~-------} 
 & Logistic Regression & 65.77 & 5.4 & \textbf{69.68} & 34.74 & 69.32 & 34.12 \\ \hhline{~-------} 
 & Naive Bayes & 50.49 & 43.04 & 52.57 & \textbf{50} & 48.9 & 46.27 \\ \hhline{~-------} 
 & SVM Linear & 66.26 & 2.12 & 68.09 & 16.08 & 68.46 & 16.23 \\ \hline

\multirow{9}{*}{4} & \multirow{3}{*}{Algorithm} & \multicolumn{6}{c|}{Word Embedding Technique} \\ \hhline{~~------} 
 &  & \multicolumn{2}{c|}{BERT} & \multicolumn{2}{c|}{Word2Vec} & \multicolumn{2}{c|}{fastText} \\ \hhline{~~------} 
 &  & \textbf{ACC (\%)} & \textbf{F1 (\%)} & \textbf{ACC (\%)} & \textbf{F1 (\%)} & \textbf{ACC (\%)} & \textbf{F1 (\%)} \\ \hhline{~-------} 
 & KNN (K = 5) & 85.33 & 76.65 & 87.41 & 80.38 & 79.46 & 64.85 \\ \hhline{~-------} 
 & Decision Tree & 85.94 & 78.66 & 82.4 & 73.63 & 82.4 & 73.63 \\ \hhline{~-------} 
 & Random Forest & \textbf{88.02} & \textbf{81.01} & 72.62 & 43.43 & 77.26 & 55.07 \\ \hhline{~-------} 
 & Logistic Regression & 81.17 & 67.78 & 84.6 & 75.29 & 83.37 & 73.44 \\ \hhline{~-------} 
 & Naive Bayes & 39.49 & 48.81 & 52.08 & 51.12 & 50.98 & 52.43 \\ \hhline{~-------} 
 & SVM Linear & 81.66 & 67.53 & 84.6 & 75.2 & 83.74 & 73.87 \\ \hline

\multirow{9}{*}{5} & \multirow{3}{*}{Algorithm} & \multicolumn{6}{c|}{Word Embedding Technique} \\ \hhline{~~------} 
 &  & \multicolumn{2}{c|}{BERT} & \multicolumn{2}{c|}{Word2Vec} & \multicolumn{2}{c|}{fastText} \\ \hhline{~~------} 
 &  & \textbf{ACC (\%)} & \textbf{F1 (\%)} & \textbf{ACC (\%)} & \textbf{F1 (\%)} & \textbf{ACC (\%)} & \textbf{F1 (\%)} \\ \hhline{~-------} 
 & KNN (K = 5) & 77.14 & 63.97 & 78.24 & 64.11 & 75.18 & 57.8 \\ \hhline{~-------} 
 & Decision Tree & 78.97 & 64.46 & 72.49 & 59.31 & 71.03 & 58.49 \\ \hhline{~-------} 
 & Random Forest & 78.48 & 64.37 & 73.84 & 44.85 & 71.52 & 40.41 \\ \hhline{~-------} 
 & Logistic Regression & 81.9 & \textbf{69.17} & 78.85 & 64.48 & 79.1 & 64.88 \\ \hhline{~-------} 
 & Naive Bayes & 39.49 & 48.92 & 51.96 & 50.81 & 50.86 & 50.25 \\ \hhline{~-------} 
 & SVM Linear & \textbf{81.91} & 68.91 & 80.81 & 66.66 & 80.32 & 65.81 \\ \hline

\multirow{9}{*}{6} & \multirow{3}{*}{Algorithm} & \multicolumn{6}{c|}{Word Embedding Technique} \\ \hhline{~~------} 
 &  & \multicolumn{2}{c|}{BERT} & \multicolumn{2}{c|}{Word2Vec} & \multicolumn{2}{c|}{fastText} \\ \hhline{~~------} 
 &  & \textbf{ACC (\%)} & \textbf{F1 (\%)} & \textbf{ACC (\%)} & \textbf{F1 (\%)} & \textbf{ACC (\%)} & \textbf{F1 (\%)} \\ \hhline{~-------} 
 & KNN (K = 5) & 41.32 & 46.78 & 63.33 & 34.78 & 63.69 & 35.29 \\ \hhline{~-------} 
 & Decision Tree & 66.63 & 2.85 & 58.68 & 40.7 & 60.64 & 44.86 \\ \hhline{~-------} 
 & Random Forest & 66.5 & 3.52 & 66.87 & 28.87 & 67.6 & 28.95 \\ \hhline{~-------} 
 & Logistic Regression & 66.5 & 2.14 & \textbf{70.41} & 34.95 & 69.8 & 34.13 \\ \hhline{~-------} 
 & Naive Bayes & 39.61 & 48.97 & 52.08 & \textbf{50.63} & 50.37 & 47.81 \\ \hhline{~-------} 
 & SVM Linear & 66.5 & 2.14 & 67.97 & 13.25 & 67.85 & 10.85 \\ \hline

\multirow{9}{*}{7} & \multirow{3}{*}{Algorithm} & \multicolumn{6}{c|}{Word Embedding Technique} \\ \hhline{~~------} 
 &  & \multicolumn{2}{c|}{BERT} & \multicolumn{2}{c|}{Word2Vec} & \multicolumn{2}{c|}{fastText} \\ \hhline{~~------} 
 &  & \textbf{ACC (\%)} & \textbf{F1 (\%)} & \textbf{ACC (\%)} & \textbf{F1 (\%)} & \textbf{ACC (\%)} & \textbf{F1 (\%)} \\ \hhline{~-------} 
 & KNN (K = 5) & 76.28 & 61.35 & \textbf{87.04} & 79.92 & 79.34 & 64.72 \\ \hhline{~-------} 
 & Decision Tree & 86.92 & \textbf{80.58} & 82.89 & 74.45 & 81.42 & 72.06 \\ \hhline{~-------} 
 & Random Forest & 80.81 & 67.09 & 72.01 & 40.83 & 72 & 40.21 \\ \hhline{~-------} 
 & Logistic Regression & 81.66 & 68.88 & 84.23 & 74.66 & 83.74 & 73.66 \\ \hhline{~-------} 
 & Naive Bayes & 49.88 & 46.34 & 54.16 & 52.71 & 48.53 & 48.22 \\ \hhline{~-------} 
 & SVM Linear & 81.54 & 67.67 & 84.35 & 75 & 83.86 & 74.12 \\ \hline

\end{tabular}
\end{table}


\begin{thebibliography}{10}
\providecommand{\url}[1]{#1}
\csname url@samestyle\endcsname
\providecommand{\newblock}{\relax}
\providecommand{\bibinfo}[2]{#2}
\providecommand{\BIBentrySTDinterwordspacing}{\spaceskip=0pt\relax}
\providecommand{\BIBentryALTinterwordstretchfactor}{4}
\providecommand{\BIBentryALTinterwordspacing}{\spaceskip=\fontdimen2\font plus
\BIBentryALTinterwordstretchfactor\fontdimen3\font minus \fontdimen4\font\relax}
\providecommand{\BIBforeignlanguage}[2]{{%
\expandafter\ifx\csname l@#1\endcsname\relax
\typeout{** WARNING: IEEEtran.bst: No hyphenation pattern has been}%
\typeout{** loaded for the language `#1'. Using the pattern for}%
\typeout{** the default language instead.}%
\else
\language=\csname l@#1\endcsname
\fi
#2}}
\providecommand{\BIBdecl}{\relax}
\BIBdecl

\bibitem{kang2018dataset}
D.~Kang, W.~Ammar, B.~Dalvi, M.~Van~Zuylen, S.~Kohlmeier, E.~Hovy, and R.~Schwartz, ``A dataset of peer reviews (peerread): Collection, insights and nlp applications,'' \emph{arXiv preprint arXiv:1804.09635}, 2018.

\bibitem{devlin2018bert}
J.~Devlin, M.-W. Chang, K.~Lee, and K.~Toutanova, ``Bert: Pre-training of deep bidirectional transformers for language understanding,'' \emph{arXiv preprint arXiv:1810.04805}, 2018.

\bibitem{dong2019unified}
L.~Dong, N.~Yang, W.~Wang, F.~Wei, X.~Liu, Y.~Wang, J.~Gao, M.~Zhou, and H.-W. Hon, ``Unified language model pre-training for natural language understanding and generation,'' \emph{Advances in Neural Information Processing Systems}, vol.~32, 2019.

\bibitem{pradhan2021deep}
T.~Pradhan, C.~Bhatia, P.~Kumar, and S.~Pal, ``A deep neural architecture based meta-review generation and final decision prediction of a scholarly article,'' \emph{Neurocomputing}, vol. 428, pp. 218--238, 2021.

\bibitem{raffel2019exploring}
C.~Raffel, N.~Shazeer, A.~Roberts, K.~Lee, S.~Narang, M.~Matena, Y.~Zhou, W.~Li, and P.~J. Liu, ``Exploring the limits of transfer learning with a unified text-to-text transformer,'' \emph{arXiv preprint arXiv:1910.10683}, 2019.

\bibitem{lin2004rouge}
C.-Y. Lin, ``Rouge: A package for automatic evaluation of summaries,'' in \emph{Text summarization branches out}, 2004, pp. 74--81.

\bibitem{wang2018sentiment}
K.~Wang and X.~Wan, ``Sentiment analysis of peer review texts for scholarly papers,'' in \emph{The 41st International ACM SIGIR Conference on Research \& Development in Information Retrieval}, 2018, pp. 175--184.

\bibitem{chakraborty2020aspect}
S.~Chakraborty, P.~Goyal, and A.~Mukherjee, ``Aspect-based sentiment analysis of scientific reviews,'' in \emph{Proceedings of the ACM/IEEE Joint Conference on Digital Libraries in 2020}, 2020, pp. 207--216.

\bibitem{ghosal2019deepsentipeer}
T.~Ghosal, R.~Verma, A.~Ekbal, and P.~Bhattacharyya, ``Deepsentipeer: Harnessing sentiment in review texts to recommend peer review decisions,'' in \emph{Proceedings of the 57th Annual Meeting of the Association for Computational Linguistics}, 2019, pp. 1120--1130.

\bibitem{kumar2022deepaspeer}
S.~Kumar, H.~Arora, T.~Ghosal, and A.~Ekbal, ``Deepaspeer: towards an aspect-level sentiment controllable framework for decision prediction from academic peer reviews,'' in \emph{Proceedings of the 22nd ACM/IEEE Joint Conference on Digital Libraries}, 2022, pp. 1--11.

\bibitem{jen2018predicting}
W.~Jen, S.~Zhang, and M.~Chen, ``Predicting conference paper acceptance,'' 2018.

\bibitem{skorikov2020machine}
M.~Skorikov and S.~Momen, ``Machine learning approach to predicting the acceptance of academic papers,'' in \emph{2020 IEEE International Conference on Industry 4.0, Artificial Intelligence, and Communications Technology (IAICT)}.\hskip 1em plus 0.5em minus 0.4em\relax IEEE, 2020, pp. 113--117.

\bibitem{bao2021predicting}
P.~Bao, W.~Hong, and X.~Li, ``Predicting paper acceptance via interpretable decision sets,'' in \emph{Companion Proceedings of the Web Conference 2021}, 2021, pp. 461--467.

\bibitem{wang2021paper}
W.~Wang, J.~Zhang, F.~Zhou, P.~Chen, and B.~Wang, ``Paper acceptance prediction at the institutional level based on the combination of individual and network features,'' \emph{Scientometrics}, vol. 126, no.~2, pp. 1581--1597, 2021.

\bibitem{wang2020reviewrobot}
Q.~Wang, Q.~Zeng, L.~Huang, K.~Knight, H.~Ji, and N.~F. Rajani, ``Reviewrobot: Explainable paper review generation based on knowledge synthesis,'' \emph{arXiv preprint arXiv:2010.06119}, 2020.

\bibitem{bharti2023peerrec}
P.~K. Bharti, T.~Ghosal, M.~Agarwal, and A.~Ekbal, ``Peerrec: An ai-based approach to automatically generate recommendations and predict decisions in peer review,'' \emph{International Journal on Digital Libraries}, pp. 1--18, 2023.

\bibitem{louabstractive}
Z.~Lou and J.~Zhang, ``Abstractive summarization on covid-19 publications.''

\bibitem{esteva2020co}
A.~Esteva, A.~Kale, R.~Paulus, K.~Hashimoto, W.~Yin, D.~Radev, and R.~Socher, ``Co-search: Covid-19 information retrieval with semantic search, question answering, and abstractive summarization,'' \emph{arXiv preprint arXiv:2006.09595}, 2020.

\bibitem{kieuvongngam2020automatic}
V.~Kieuvongngam, B.~Tan, and Y.~Niu, ``Automatic text summarization of covid-19 medical research articles using bert and gpt-2,'' \emph{arXiv preprint arXiv:2006.01997}, 2020.

\bibitem{liu2018generative}
L.~Liu, Y.~Lu, M.~Yang, Q.~Qu, J.~Zhu, and H.~Li, ``Generative adversarial network for abstractive text summarization,'' in \emph{Proceedings of the AAAI Conference on Artificial Intelligence}, vol.~32, no.~1, 2018.

\bibitem{li2017deep}
P.~Li, W.~Lam, L.~Bing, and Z.~Wang, ``Deep recurrent generative decoder for abstractive text summarization,'' \emph{arXiv preprint arXiv:1708.00625}, 2017.

\bibitem{yuan2021can}
W.~Yuan, P.~Liu, and G.~Neubig, ``Can we automate scientific reviewing?'' \emph{arXiv preprint arXiv:2102.00176}, 2021.

\bibitem{kang2018a}
D.~{Kang}, W.~{Ammar}, B.~{Dalvi}, M.~van {Zuylen}, S.~{Kohlmeier}, E.~H. {Hovy}, and R.~{Schwartz}, ``A dataset of peer reviews (peerread): Collection, insights and nlp applications,'' in \emph{Proceedings of the 2018 Conference of the North American Chapter of the Association for Computational Linguistics: Human Language Technologies, Volume 1 (Long Papers)}, vol.~1, 2018, pp. 1647--1661.

\bibitem{kumar2023towards}
A.~Kumar, T.~Ghosal, S.~Bhattacharjee, and A.~Ekbal, ``Towards automated meta-review generation via an nlp/ml pipeline in different stages of the scholarly peer review process,'' \emph{International Journal on Digital Libraries}, pp. 1--12, 2023.

\bibitem{kumar2023deepmetagen}
S.~Kumar, T.~Ghosal, and A.~Ekbal, ``Deepmetagen: an unsupervised deep neural approach to generate template-based meta-reviews leveraging on aspect category and sentiment analysis from peer reviews,'' \emph{International Journal on Digital Libraries}, vol.~24, no.~4, pp. 263--281, 2023.

\bibitem{kousha2024artificial}
K.~Kousha and M.~Thelwall, ``Artificial intelligence to support publishing and peer review: A summary and review,'' \emph{Learned Publishing}, vol.~37, no.~1, pp. 4--12, 2024.

\bibitem{hasan2023review}
M.~T. Hasan, M.~A.~E. Hossain, M.~S.~H. Mukta, A.~Akter, M.~Ahmed, and S.~Islam, ``A review on deep-learning-based cyberbullying detection,'' \emph{Future Internet}, vol.~15, no.~5, p. 179, 2023.

\bibitem{ramos2003using}
J.~Ramos \emph{et~al.}, ``Using tf-idf to determine word relevance in document queries,'' in \emph{Proceedings of the first instructional conference on machine learning}, vol. 242, no.~1.\hskip 1em plus 0.5em minus 0.4em\relax Citeseer, 2003, pp. 29--48.

\bibitem{mikolov2013efficient}
T.~Mikolov, K.~Chen, G.~Corrado, and J.~Dean, ``Efficient estimation of word representations in vector space,'' \emph{arXiv preprint arXiv:1301.3781}, 2013.

\bibitem{mikolov2013distributed}
T.~Mikolov, I.~Sutskever, K.~Chen, G.~S. Corrado, and J.~Dean, ``Distributed representations of words and phrases and their compositionality,'' in \emph{Advances in neural information processing systems}, 2013, pp. 3111--3119.

\bibitem{pennington2014glove}
J.~Pennington, R.~Socher, and C.~D. Manning, ``Glove: Global vectors for word representation,'' in \emph{Proceedings of the 2014 conference on empirical methods in natural language processing (EMNLP)}, 2014, pp. 1532--1543.

\bibitem{peters2018deep}
M.~E. Peters, M.~Neumann, M.~Iyyer, M.~Gardner, C.~Clark, K.~Lee, and L.~Zettlemoyer, ``Deep contextualized word representations,'' \emph{arXiv preprint arXiv:1802.05365}, 2018.

\bibitem{joulin2016bag}
A.~Joulin, E.~Grave, P.~Bojanowski, and T.~Mikolov, ``Bag of tricks for efficient text classification,'' \emph{arXiv preprint arXiv:1607.01759}, 2016.

\bibitem{fix1989discriminatory}
E.~Fix and J.~L. Hodges, ``Discriminatory analysis. nonparametric discrimination: Consistency properties,'' \emph{International Statistical Review/Revue Internationale de Statistique}, vol.~57, no.~3, pp. 238--247, 1989.

\bibitem{altman1992introduction}
N.~S. Altman, ``An introduction to kernel and nearest-neighbor nonparametric regression,'' \emph{The American Statistician}, vol.~46, no.~3, pp. 175--185, 1992.

\bibitem{quinlan1987simplifying}
J.~R. Quinlan, ``Simplifying decision trees,'' \emph{International journal of man-machine studies}, vol.~27, no.~3, pp. 221--234, 1987.

\bibitem{breiman2001random}
L.~Breiman, ``Random forests,'' \emph{Machine learning}, vol.~45, no.~1, pp. 5--32, 2001.

\bibitem{tranmer2008binary}
M.~Tranmer and M.~Elliot, ``Binary logistic regression,'' \emph{Cathie Marsh for census and survey research, paper}, vol.~20, 2008.

\bibitem{huang2011naive}
Y.~Huang and L.~Li, ``Naive bayes classification algorithm based on small sample set,'' in \emph{2011 IEEE International conference on cloud computing and intelligence systems}.\hskip 1em plus 0.5em minus 0.4em\relax IEEE, 2011, pp. 34--39.

\bibitem{cortes1995support}
C.~Cortes and V.~Vapnik, ``Support-vector networks,'' \emph{Machine learning}, vol.~20, no.~3, pp. 273--297, 1995.

\bibitem{mihalcea2004textrank}
R.~Mihalcea and P.~Tarau, ``Textrank: Bringing order into text,'' in \emph{Proceedings of the 2004 conference on empirical methods in natural language processing}, 2004, pp. 404--411.

\bibitem{radford2019language}
A.~Radford, J.~Wu, R.~Child, D.~Luan, D.~Amodei, I.~Sutskever \emph{et~al.}, ``Language models are unsupervised multitask learners,'' \emph{OpenAI blog}, vol.~1, no.~8, p.~9, 2019.

\bibitem{lewis2019bart}
M.~Lewis, Y.~Liu, N.~Goyal, M.~Ghazvininejad, A.~Mohamed, O.~Levy, V.~Stoyanov, and L.~Zettlemoyer, ``Bart: Denoising sequence-to-sequence pre-training for natural language generation, translation, and comprehension,'' \emph{arXiv preprint arXiv:1910.13461}, 2019.

\bibitem{zhang2020pegasus}
J.~Zhang, Y.~Zhao, M.~Saleh, and P.~Liu, ``Pegasus: Pre-training with extracted gap-sentences for abstractive summarization,'' in \emph{International Conference on Machine Learning}.\hskip 1em plus 0.5em minus 0.4em\relax PMLR, 2020, pp. 11\,328--11\,339.

\bibitem{tay2019red}
W.~Tay, A.~Joshi, X.~J. Zhang, S.~Karimi, and S.~Wan, ``Red-faced rouge: Examining the suitability of rouge for opinion summary evaluation,'' in \emph{Proceedings of the The 17th Annual Workshop of the Australasian Language Technology Association}, 2019, pp. 52--60.

\bibitem{lin2004automatic}
C.-Y. Lin and F.~J. Och, ``Automatic evaluation of machine translation quality using longest common subsequence and skip-bigram statistics,'' in \emph{Proceedings of the 42nd Annual Meeting of the Association for Computational Linguistics (ACL-04)}, 2004, pp. 605--612.

\bibitem{sanchez2020experimental}
J.~M. Sanchez-Gomez, M.~A. Vega-Rodr{\'\i}guez, and C.~J. P{\'e}rez, ``Experimental analysis of multiple criteria for extractive multi-document text summarization,'' \emph{Expert Systems with Applications}, vol. 140, p. 112904, 2020.

\bibitem{woolson2007wilcoxon}
R.~F. Woolson, ``Wilcoxon signed-rank test,'' \emph{Wiley encyclopedia of clinical trials}, pp. 1--3, 2007.

\bibitem{mah2023art}
P.~M. Mah, ``The art of deep learning and natural language processing for emotional sentiment analysis on the academic scholars' peer review process.'' in \emph{Proceedings of the 24th Annual Conference on Information Technology Education}, 2023, pp. 186--198.

\bibitem{meng2023assessing}
J.~Meng, ``Assessing and predicting the quality of peer reviews: a text mining approach,'' \emph{The Electronic Library}, vol.~41, no. 2/3, pp. 186--203, 2023.

\end{thebibliography}
\end{document}